\documentclass{article}

\usepackage{microtype}
\usepackage{graphicx}
\usepackage{subcaption}
\usepackage{booktabs} 
\usepackage{color}

\usepackage[utf8]{inputenc} 
\usepackage[T1]{fontenc}    

\usepackage{url}            
\usepackage{amsfonts}       
\usepackage{nicefrac}       
\usepackage{fixmath}

\usepackage[acronym,smallcaps,nowarn]{glossaries}
\glsdisablehyper
\makeglossaries

\usepackage[usenames,dvipsnames]{xcolor}
\definecolor{shadecolor}{gray}{0.9}

\usepackage{booktabs, array}

\newacronym{MAP}{map}{maximum a posteriori}
\newacronym{MLE}{mle}{maximum likelihood estimation}

\newacronym{MLP}{mlp}{multilayer perceptron}
\newacronym{RELU}{ReLU}{rectified linear unit}

\newacronym[longplural={Gaussian Processes}]{GP}{gp}{Gaussian Process}

\newacronym{ARD}{ard}{automatic relevance determination}

\newacronym{BO}{bo}{Bayesian optimization}
\newacronym{HPO}{hpo}{hyperparameter optimization}

\newacronym{MISO}{miso}{Multi-Information Source Optimization}

\newacronym{FABOLAS}{fabolas}{FAst Bayesian Optimization on LArge data Sets}

\newacronym{HB}{hb}{Hyperband}

\newacronym{MTBO}{mtbo}{Multi-Task Bayesian Optimization}

\newcommand{\nestedmathbold}[1]{{\mathbold{#1}}}

\newcommand{\mbp}{\nestedmathbold{p}}

\newcommand{\mbx}{\nestedmathbold{x}}
\newcommand{\mby}{\nestedmathbold{y}}

\newcommand{\mbI}{\nestedmathbold{I}}

\newcommand{\mbK}{\nestedmathbold{K}}
\newcommand{\mbL}{\nestedmathbold{L}}

\newcommand{\mbP}{\nestedmathbold{P}}

\newcommand{\mbX}{\nestedmathbold{X}}
\newcommand{\mbY}{\nestedmathbold{Y}}

\DeclareMathOperator*{\argmin}{arg\,min}

\newcommand{\cD}{\mathcal{D}}

\newcommand{\cN}{\mathcal{N}}

\newcommand{\cX}{\mathcal{X}}

\newcommand{\bbR}{\mathbb{R}}

\usepackage{wrapfig}
\usepackage{pgfplots}

\PassOptionsToPackage{numbers, compress}{natbib}

\usepackage[preprint]{neurips_2020}

\usepackage[utf8]{inputenc} 
\usepackage[T1]{fontenc}    
\usepackage{hyperref}       
\usepackage{url}            
\usepackage{booktabs}       
\usepackage{amsfonts}       
\usepackage{nicefrac}       
\usepackage{microtype}      

\DeclareRobustCommand{\parhead}[1]{\textbf{#1}~}

\newcommand{\figref}[1]{Figure~\ref{fig:#1}}

\newcommand{\secref}[1]{Section~\ref{sec:#1}}
\renewcommand{\eqref}[1]{Eq.~\ref{eq:#1}}
\newcommand{\eqp}[1]{(\ref{eq:#1})}

\newcommand{\appref}[1]{Appendix~\ref{app:#1}}
\mathchardef\mhyphen="2D

\newif\ifappendix
\appendixtrue  

\title{Model-based Asynchronous Hyperparameter and Neural Architecture Search}

\author{
Aaron Klein{$^*$}\\
Amazon Web Services\\
\texttt{kleiaaro@amazon.com} \\ 
\And Louis Tiao\thanks{These authors contributed equally.}\\
University of Sydney\\
\texttt{louis.tiao@sydney.edu.au} \\
\And Thibaut Lienart\\
Amazon Web Services\\
\texttt{tlienart@amazon.com} \\ 
\And C\'{e}dric Archambeau\\
Amazon Web Services\\
\texttt{cedrica@amazon.com} \\ 
\And Matthias Seeger \\
Amazon Web Services\\
\texttt{matthis@amazon.com} \\ 
}

\begin{document}

\maketitle

\begin{abstract}
We introduce a model-based asynchronous multi-fidelity method for hyperparameter and neural architecture search that combines the strengths of asynchronous Hyperband and Gaussian process-based Bayesian optimization. At the heart of our method is a probabilistic model that can simultaneously reason across hyperparameters and resource levels, and supports decision-making in the presence of pending evaluations. We demonstrate the effectiveness of our method on a wide range of challenging benchmarks, for tabular data, image classification and language modelling, and report substantial speed-ups over current state-of-the-art methods. Our new methods, along with asynchronous baselines, are implemented in a distributed framework which will be open sourced along with this publication.
\end{abstract}

\section{Introduction}
\label{sec:introduction}

The goal of hyperparameter and neural architecture search (HNAS) is to automate the process of finding the architecture or hyperparameters $\mbx_{\star} \in \argmin_{\mbx \in \cX} f(\mbx)$ for a deep learning algorithm which minimizes a validation loss $f(\mbx)$, observed through noise:
$y_i = f(\mbx_i) + \epsilon_i,\quad \epsilon_i \sim \cN(0, \sigma^2), i=1,\dots, n$.
Bayesian optimization (BO)~\citep{jones-jgo98a,shahriari-ieee16a} is an effective model-based approach for these expensive optimization problems. It constructs a probabilistic surrogate model of the loss function $p(f \mid \cD)$ based on previous evaluations $\cD = \{(\mbx_i, y_i)\}_{i=1}^n$. Searching for the global minimum of $f$ is then driven by trading off exploration in regions where our model is uncertain and exploitation in regions where the global optimum is assumed to be located.

\begin{figure}[ht]
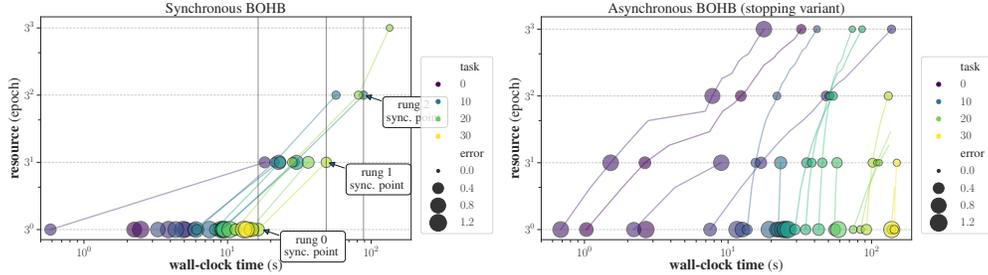

  \centering
  \begin{subfigure}[t]{0.47\columnwidth}
    \resizebox{\columnwidth}{!}{\input{plots/hyperband_synchronous.pgf}}
  \end{subfigure}
  \begin{subfigure}[t]{0.47\columnwidth}
    \resizebox{\columnwidth}{!}{\input{plots/hyperband_stopping.pgf}}
  \end{subfigure}
 
\caption{Evolution of synchronous (left) and asynchronous (right) BOHB on OpenML MLP tuning (see \secref{experiments}). x: wall-clock time (150 secs), y: training epochs done (log scale). Coloured lines trace training jobs for HP configurations, with marker size proportional to validation error (smaller is better). Lines not reaching the maximum number of authorized epochs (27) are stopped early. For synchronous BOHB, rungs (epochs 1, 3, 9) need to be filled completely before any configuration can be promoted, and each synchronization results in some idle time. In asychronous BOHB, configurations proceed to higher epochs faster. Without synchronization overhead, more configurations are promoted to the maximum number of epoch in the same wall-clock time (figure best seen in colours).}
  \label{fig:teasers}
\end{figure}

For HNAS problems, standard BO needs to be augmented in order to remain competitive. Training runs for unpromising configurations $\mbx$ can be stopped early, and thereby serve as low-fidelity approximations of $f(\mbx)$~\citep{Swersky:14,domhan-ijcai15,klein-iclr17}. Also, evaluations of $f$ can be executed in parallel to reduce the wall-clock time required for finding a good solution. Successive halving (SH)~\citep{jamieson-aistats16} and Hyperband (HB)~\cite{li-iclr17} are simple, easily parallelizable multi-fidelity scheduling algorithms. 
BOHB~\citep{falkner-icml18} combines the efficient any-time performance of HB with a probabilistic model to guide the search. However, both HB and BOHB are {\em synchronous} algorithms: stopping decisions are done only after synchronizing all training jobs at certain resource levels (called {\em rungs}). This can be wasteful when the evaluation of some configurations take longer than others, as it is often the case for HNAS~\citep{ying-arxiv19}, and can substantially delay progress for high-performing configurations (see \figref{teasers} for an illustration). Asynchronous parallel variants of HB have been proposed \citep{li-arxiv18}, but rely on a uniform random sampling of architectures and hyperparameters.

\subsection{Contributions}

Dealing with HPO in general and HNAS for neural networks, we would like to make the most efficient use of a given parallel computation budget (e.g., number of GPU cores for a certain amount of time) in order to find high-accuracy solutions in the least amount of wall-clock time. Besides exploiting low-fidelity approximations, we demonstrate that {\em asynchronous parallel} scheduling is a key property for cost-efficient search. Our novel combination of multi-fidelity BO with asynchronous parallel scheduling, dubbed asynchronous BOHB (A-BOHB), can substantially outperform either of them in isolation. More specifically:
\begin{itemize}
	\item We clarify differences between existing asynchronous HB extensions. In our experiments, ASHA, as described by Li et al.~\citep{li-arxiv18} is frequently outperformed by a simpler stopping rule variant, related to the median rule \citep{golovin-kdd17} and first implemented in Ray Tune \citep{liaw2018tune}.
\item We introduce a new model-based extension of asynchronous HB which handles multi-fidelity evaluations in a principled way, simultaneously reasoning across resource levels and asynchronously across workers. The underlying surrogate model is a joint Gaussian process (GP) akin the ones used in \citep{Swersky:13, Swersky:14}.
\item By sampling from our model, we often achieve the same performance in the same amount of wall-clock time but with half the computational resources as sampling uniformly at random.
\item On neural network benchmarks, we show that A-BOHB is more efficient in terms of wall-clock time than other state-of-the-art algorithms, especially if evaluations are expensive. Synchronous BOHB \citep{falkner-icml18} suffers from substantial synchronization overheads, while variants of asynchronous HB fail to approach the global optima.
\end{itemize}

Next, we relate our work to approaches recently published in the literature. In \secref{schedule}, we review synchronous SH and HB, as well as two asynchronous extensions. Our novel method is presented in \secref{model}. We present empirical evaluations for the HNAS of various neural architecture types in \secref{experiments}, and finish with conclusions and future work in \secref{conclusions}.

\subsection{Related Work}\label{sec:related_work}

A range of prior Gaussian process (GP) based BO work exploits multiple fidelities of the objective~\citep{Kennedy:00,Klein:17,Poloczek:18,Cutajar:18}. A joint GP model across configurations and tasks was used by Swersky et al. \cite{Swersky:13}, allowing for trade-offs between cheap auxiliaries and the expensive target. Klein et al.~\citep{Klein:17} presented a continuous multi-fidelity BO method, where training set size is an input. It relies on a complex and expensive acquisition function, and experiments with asynchronous scheduling are not provided. The ``freeze-thaw'' approach \citep{Swersky:14} allows for asynchronous parallel evaluations, yet its scheduling is quite different from asynchronous HB. They propose an exponential decay kernel, which we use here as well. No results for deep neural network tuning were presented, and no implementation is publicly available. Previous work on asynchronous BO has demonstrated performance gains over synchronous BO methods~\citep{alvin-icml19, kandasamy-aistats18}, yet this work did not exploit multiple fidelities of the objective.

Synchronous HB was combined with BO in the BOHB method~\citep{falkner-icml18}, using TPE~\citep{Bergstra:11} for BO. It combines early speed-ups of HB with fast convergence later on, and shows competitive performance for NAS \citep{ying-arxiv19, dong-arXiv20}. BOHB fits independent surrogate models at each resource level, which hinders learning curve interpolation. Being based on TPE, it is unclear how to extend BOHB to the asynchronous setting.

\section{Synchronous and Asynchronous Multi-Fidelity Scheduling}
\label{sec:schedule}

Multi-fidelity HNAS considers an objective $f(\mbx, r)$, indexed by the {\em resource level} $r\in \{r_{\text{min}}, \dotsc, r_{\text{max}} \}$. The target function of interest is $f(\mbx) = f(\mbx, r_{\text{max}})$, while evaluations of $f(\mbx, r)$ for $r < r_{\text{max}}$ are lower-fidelity evaluations that may be correlated with $f(\mbx)$ and therefore help guide the search. Here, we consider the number of training epochs ({\it i.e.}, full sweeps over the data) as the resource $r$, though other choices can be sensible as well ({\it e.g.}, training subset ratio\citep{Klein:17}).

Parallel multi-fidelity HNAS methods need to make two kinds of decisions: {\em choosing} configurations to evaluate and {\em scheduling} their evaluations (including resource allocation). Scheduling decisions happen at certain resource {\em rungs}, chosen according to the SH principle~\cite{jamieson-aistats16}. Let $\eta\in\{2, 3, 4\}$ be the halving constant, $r_{\text{max}}$ and $r_{\text{min}}$ the maximum and minimum resource level for an evaluation (assuming, for simplicity, that $r_{\text{max}} / r_{\text{min}} = \eta^K$). The full set of rungs is $\mathcal{R} = \{ r_{\text{min}}\eta^k\; |\; k=0,\dots, K \}$. Then, {\em brackets} are (ordered) subsets $B_s = \{ r_{\text{min}}\eta^{k+s}\; |\; k=0,\dots, K-s \} \subseteq \mathcal{R}$, where $s=0,\dots, K$. In the sequel, the term ``rung'' will be overloaded to both mean the resource level at which stop/go decisions are made, and the list of configs (with metric values) which reached this level.

We distinguish between {\em synchronous} and {\em asynchronous} scheduling. The former is detailed in \citep{li-iclr17,falkner-icml18}. In a nutshell, each rung of each bracket has an \textit{a priori} fixed size ({\it i.e.}, number of slots for configs evaluated until the rung level), the size ratio between successive rungs being $\eta^{-1}$. Hyperparameter configurations promoted to a rung are evaluated until the corresponding resource level, making use of parallel computation. Once all of them finish, the $\eta^{-1}$ fraction of top performing configurations are promoted to the next rung while the others are terminated. Each rung is a synchronization point for many training tasks: it has to be populated entirely before any configuration can be promoted. In an outer loop, the HB algorithm cycles over brackets $s=0,\dots, K$ periodically until the total budget is spent. The original SH algorithm is obtained as a special case, using only the single bracket $B_0$.

We provide a unified perspective on two different asynchronous generalizations of HB, proposed in prior work. For any hyperparameter and architecture configuration, its bracket index $s\in\{0,\dots, K\}$ determines the rung levels $B_s$ at which scheduling decisions are made, sampled from a distribution $P(s)$. Our choice of $P(s)$ is provided in \ifappendix \appref{bracket_distribution}. \else the supplemental material. \fi
Once an evaluation of $\mbx$ reaches a rung $r\in B_s$, its metric value $y$ is recorded there. Then, the binary predicate $\texttt{continue}_s(\mbx, r, y)$ evaluates to true iff $y$ is in the top $\eta^{-1}$ fraction of records at the rung.

\parhead{Stopping variant.}
This is a simple extension of the median stopping rule \citep{golovin-kdd17}, first implemented in \href{https://ray.readthedocs.io/en/latest/tune.html}{Ray Tune} \citep{liaw2018tune}. Once a job for $\mbx$ reaches a rung $r$, if $\texttt{continue}_s(\mbx, r, y)$ is true, it continues towards the next rung level. Otherwise, it is {\em stopped}, and the worker becomes free to pick up a novel evaluation. As long as fewer than $\eta$ metrics are recorded at a rung, the job is always continued.

\parhead{Promotion variant (ASHA).}
This variant of asynchronous HB was presented by Li et al. \citep{li-arxiv18} and called {\em ASHA}. Once a job for some $\mbx$ reaches a rung $r$, it is {\em paused} there, and the worker is released. The evaluation of $\mbx$ can be {\em promoted} ({\it i.e.}, continued to the next rung) later on. When a worker becomes available, we sample a bracket $s$ and scan the rungs $r\in B_s$ in descending order, running a job to promote the first paused $\mbx$ for which $\texttt{continue}_s(\mbx, r, y)$ is true. When no such paused configuration exists, we start a new evaluation from scratch. With fewer than $\eta$ metrics recorded at a rung, no promotions happen there. The promotion variant needs pause-and-resume training. \footnote{
   Supporting pause-and-resume by checkpointing is difficult in the multi-node distributed context. It is simpler to start training from scratch whenever a configuration is promoted. This is done in our implementation.}

Synchronous scheduling makes less efficient use of parallel computation. Configurations in the same rung may require quite different compute times, for example, if we search over hyperparameters which control network size. At high rungs, few configurations run for much longer, and some workers may just be idle at a synchronization point (such points do not exist in the asynchronous case). Finally, rung levels have fixed sizes, which have to be filled before any configuration can be promoted, while promotions to larger resource levels can happen earlier with asynchronous scheduling. Of course, the latter risks promoting mediocre configurations simply because they were selected earlier. \figref{teasers} illustrates these differences. The stopping and promotion variants can exhibit rather different behavior. Initially, the stopping variant gives most configurations the benefit of doubt, while promotion pauses evaluations until they can be compared against a sufficient number of alternative configurations.

\section{Model-based Asynchronous Multi-Fidelity Optimization}
\label{sec:model}

We now describe our new model-based method that is able to make asynchronous decisions across different fidelities of the objective function.
The configurations to be evaluated in HB are drawn at random from the configuration space. Motivated by the fact that in standard sequential settings, Gaussian process (GP) based BO tends to outperform random search~\citep{shahriari-ieee16a}, we aim to equip asynchronous HB with decisions based on a {\em GP surrogate model}. Recall that $\mbx$ denotes the hyperparameter configuration, $r$ the resource level (see \secref{schedule}). Following a range of prior work on non-parallel multi-fidelity BO \cite{Swersky:13, Klein:17, Poloczek:18, Swersky:14}, we place a GP prior over $f(\mbx, r)$ to jointly model not only the auto-correlation within each fidelity, but also the cross-correlations between fidelities,
\begin{equation*}
  f(\mbx, r) \sim \mathcal{GP}(\mu(\mbx, r), k((\mbx, r), (\mbx', r'))).
\end{equation*}
Compared to synchronous BOHB~\citep{falkner-icml18}, which  employs an independent TPE~\citep{bergstra-nips11a} model over $\mbx$ for each fidelity $r$, our method uses a joint GP model that explicitly captures correlations across all fidelities.

Since multiple configurations are evaluated in parallel, we have to deal with {\em pending} evaluations when proposing new configurations to be evaluated. We do so by {\em fantasizing} their outcomes, as detailed by Snoek et al.~\citep{Snoek:12}. In a nutshell, we marginalize the acquisition function over the GP posterior predictive of evaluation outcomes for pending configurations. In practice, we approximate this by averaging the acquisition function over sampled function values.
It is worth noting that, as long as kernel hyperparameters remain constant, fantasizing comes almost for free. For GP regression, the posterior covariance matrix does not depend on the function values (only on the inputs), which means that fantasizing does not require additional cubic-complexity computations. Kernel hyperparameters may be updated with empirical Bayes or Markov chain Monte Carlo~\citep{Snoek:12}.

{\bf How do pending evaluations and fantasizing generalize to the asynchronous case?} The relevant covariates are now the pairs $(\mbx, r)$. As a task for $\mbx$ runs down a bracket $s$, it emits target values at $r\in B_s$ (unless stopped). The following logic is task-local and ensures that each labeled covariate was previously pending. If a task to evaluate $\mbx$ is started in bracket $s$, we register a pending evaluation for $(\mbx, r_{\text{min}}\eta^s)$. Once it reaches the next rung $r\in B_s$, we convert the \emph{pending} to a \emph{labeled} evaluation $(\mbx, r)$. Then, as detailed above, we make a stopping decision. If the configuration survives, we register $(\mbx, r \eta)$ as \emph{pending}, as $r \eta$ is the next rung it will reach. Details about how posterior distributions over both labeled and labeled plus fantasized evaluations are represented, are given in \ifappendix \appref{posterior}. \else the supplemental material. \fi All linear algebra computations are delayed until the posterior is needed for the next decision.

{\bf Choice of the kernel.} We implemented and compared a range of different surrogate models for $f(\mbx, r)$: a joint Mat\'{e}rn $\frac{5}2$  kernel with automatic relevance determination (ARD) \cite{Snoek:12} over $(\mbx, r)$, with and without warping of $r$; the additive exponential decay model of \citep{Swersky:14}; a novel non-additive extension of the latter, relaxing underlying assumptions which may not hold in practice. In all cases, the base kernel on $\mbx$ alone is Mat\'{e}rn $\frac{5}2$ ARD. Details for our new surrogate model are provided in \ifappendix \appref{surrogate}. \else the supplemental material. \fi While A-BOHB can significantly outperform synchronous BOHB or non-model-based variants, we found that the choice of joint surrogate model did not have a consistent impact on these results.

{\bf Acquisition strategy.} As in synchronous BOHB~\cite{falkner-icml18}, we employ the GP surrogate model only to choose configurations $\mbx$ for new tasks, while scheduling decisions are done as in HB. Compared to standard BO without multiple fidelities, it is much less clear how a non-myopically optimal decision would look like in this setting. In this paper, we follow synchronous BOHB in choosing $\mbx$ by optimizing the expected improvement (EI) acquisition function \cite{shahriari-ieee16a} over the posterior on $f(\cdot, r_{\text{acq}})$, where $r_{\text{acq}}$ is fixed for each decision. We choose $r_{\text{acq}}\in \mathcal{R}$ as the largest rung for which at least $L$ metric values have been observed, where $L$ is typically set to the input dimensionality (\emph{i.e.} number of hyperparameters). After some initial period, $r_{\text{acq}}$ eventually converges to $r_{\text{max}}$. This strategy does not take into account that the choice of $\mbx$ may have different downstream stopping or promotion consequences. Exploring alternative strategies is a direction left for future work.

\section{Experiments}\label{sec:experiments}

We compared our asynchronous BOHB (A-BOHB) to a range of state-of-the-art algorithms. For synchronous HB and BOHB, we used the {\tt HpBandSter} package.\footnote{\url{https://github.com/automl/HpBandSter}}
The implementation by Falkner et al. \citep{falkner-icml18} uses a simple mechanism to reduce the amount of idling workers: whenever evaluations for a rung are started, if there are more free workers than slots in the rung, left-over workers are used to immediately start the next bracket. However, this requires a logic associating workers with brackets, and still needs synchronization between workers.

We implemented A-BOHB, as well as all other asynchronous techniques, in {\tt AutoGluon}.\footnote{\url{https://github.com/awslabs/autogluon}}
A number of worker nodes run training jobs as scheduled, publishing metrics after each epoch. One of the workers doubles as master, concurrently running the HP optimization and scheduling. The master node communicates with other workers over the network. All experiments were conducted on \textsc{AWS EC2}, using \texttt{m4.xlarge} instances for CPU, and \texttt{g4dn.xlarge} instances for GPU. Our asynchronously distributed HNAS system will be open sourced in {\tt AutoGluon} along with this publication.

In all methods using SH scheduling, the halving rate was $\eta=3$. Our performance metric after wall-clock time $t$ is the {\em immediate regret} $r_t = y_t - y_{\star}$, where $y_{\star}$ is the best observed validation error across all methods, runs and time steps on this dataset. Using the regret in place of just the validation error gives a more detailed picture of progress towards the global optimum. We report mean and the standard error of the mean across multiple runs for each method. We are not aware of a publicly available implementation of Swersky et al. \cite{Swersky:14} (whose complexity is considerable), nor of follow-up work, which is why we could not compare against this technique in our experiments.

We run experiments on the following set of benchmarks. A detailed list of all hyperparameters with their corresponding ranges, along with further details, can be found in \ifappendix \appref{hpranges}.\else the supplemental material.\fi

{\bf Featurized OpenML Datasets:} For our first benchmark, we optimize the hyperparameters of a multi-layer perceptron (MLP) on tabular classification datasets gathered from OpenML~\citep{vanschoren-sigkdd13a}. Our MLP has two hidden layers with ReLU activations. We tuned 8 hyperparameters: the learning rate of ADAM~\citep{kingma-iclr15}, the batch size, and for each layer the dropout rate, number of units and the scale of a uniform weight initialization distribution. The number of epochs $r$ varies between $r_{\text{min}} = 1$ and $r_{\text{max}} = 27$, and HB is using 4 brackets. Results for each method are averaged over 30 independent runs with different random seeds. Due to space constraints, we only report results on \textit{electricity}, all other results are provided in \ifappendix \appref{results}. \else the supplemental material. \fi 

{\bf Neural Architecture Search on Tabular Benchmark:} To evaluate our approach in a NAS setting, we used the NASBench201 datasets from Dong et al. \citep{dong-arXiv20} which consists of a offline evaluated grid of neural network architectures for three different datasets: CIFAR-10, CIFAR-100~\citep{krizhevsky-tech09a} and Imagenet16~\citep{chrabaszcz-arXiv17}. Categorical hyperparameters were one-hot encoded. The minimum and maximum number of epochs were $r_{\text{min}} = 1$ and $r_{\text{max}} = 200$, and HB uses 5 brackets. To simulate realistic dynamics of asynchronous NAS, we put each worker to sleep for the time per epoch each real evaluation took. This does not speed up experiments, but decreases cost, since no GPU instances are required.

{\bf Language Modelling:} Next, we tune an LSTM~\citep{hochreiter-neural97}, applied to language modelling on the WikiText-2 dataset~\cite{merity-arXiv17} with the default training/validation/test split. We optimized the initial learning rate of SGD, batch size, dropout rate, gradient clipping range, and decay factor for the learning rate which is applied if the validation error has not improved upon the previous epoch. The LSTM consisted of two layers with 1500 hidden units each. For the word embeddings, we also used 1500 hidden units and tied them with the output. The number of epochs $r$ runs between $r_{\text{min}}=1$ and $r_{\text{max}}=81$, and HB uses 5 brackets.

{\bf Image Classification:} Lastly, we tune ResNet~\citep{he-cvpr16} on the image classification dataset CIFAR-10~\citep{krizhevsky-tech09a}, tuning the batch size, weight decay, initial learning rate and momentum. We used 40000 images for training, 10000 for validation, and applied standard data augmentation (cropping; horizontal flips). SGD is used for training, the learning rate is multiplied by $0.1$ after 25 epochs. We used $r_{\text{min}}=1$ and $r_{\text{max}}=27$, and 4 brackets for HB. As our goal is to differentiate between the methods, we neglected further tricks of the trade needed to get to state-of-the-art accuracy~\citep{yang-iclr20}.

\subsection{Comparing Random- and Model-based Approaches, and their Schedulers}

\begin{figure*}[t]
\begin{center}
 \includegraphics[width=.49\textwidth]{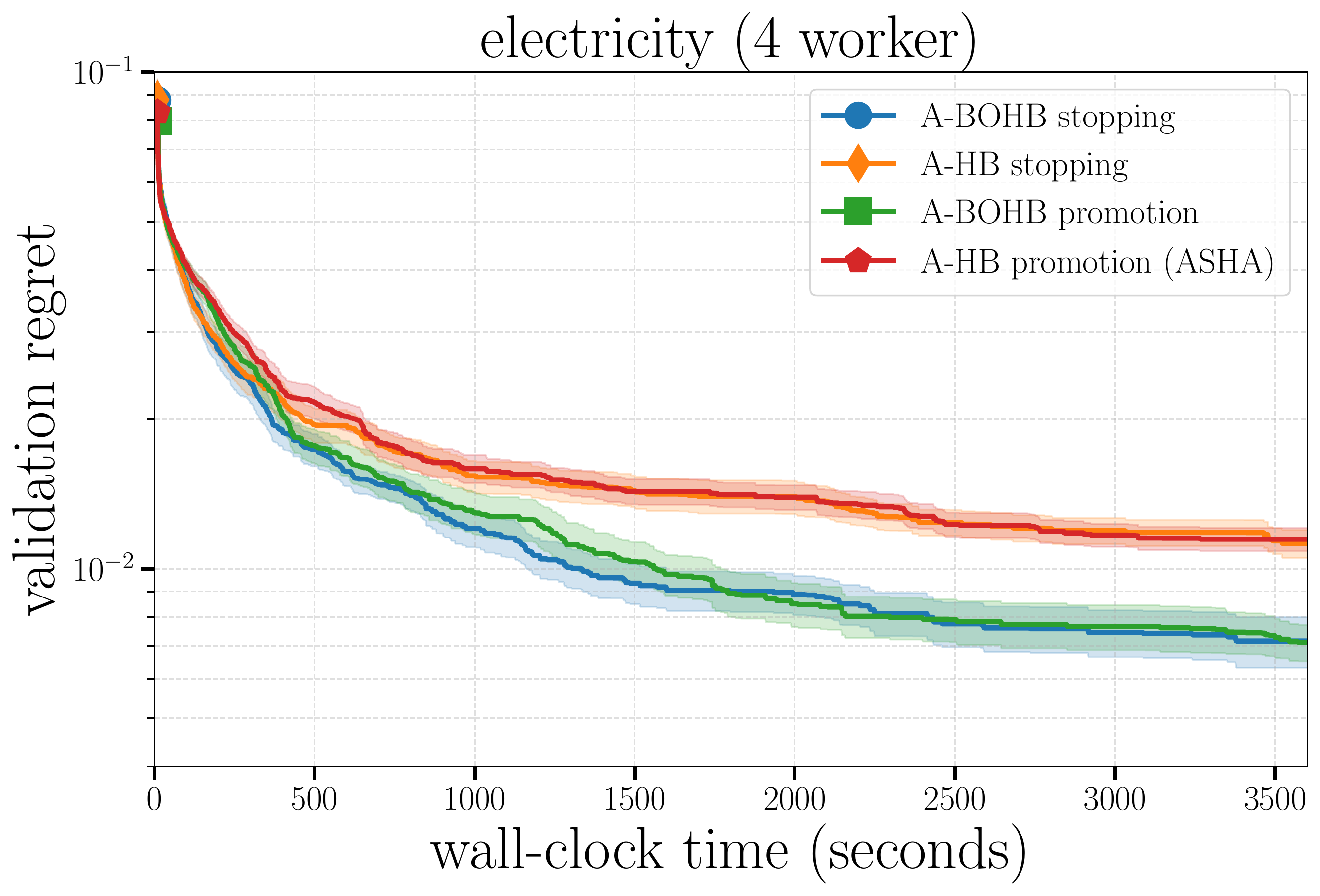}
 \includegraphics[width=.49\linewidth]{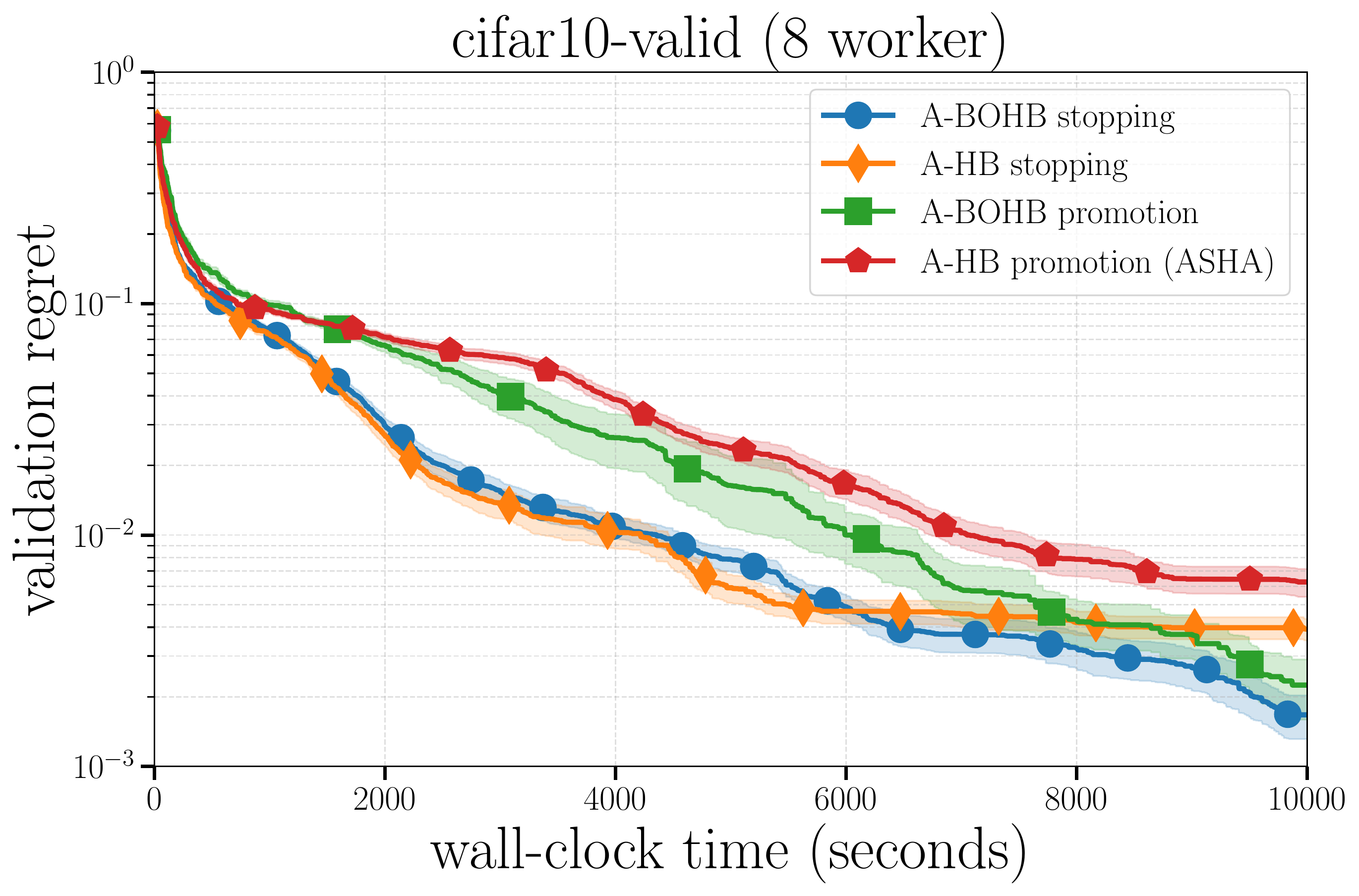}

\caption{Comparison of asynchronous HB and BOHB (stopping and promotion scheduling variants). Left: MLP on {\em electricity} (4 workers). Right: NASBench201 on CIFAR-10 (8 workers). A-BOHB clearly outperforms A-HB. Differences between the scheduling variants are insignificant for MLP, and more pronounced for NAS.}
\label{fig:async_methods}
\end{center}
\end{figure*}

We start with comparing methods using asynchronous scheduling: our BOHB (A-BOHB) and Hyperband (A-HB). Recall that scheduling comes in two variants: stopping and promotion, and that A-HB promotion is known as ASHA~\cite{li-arxiv18}. \figref{async_methods} shows the immediate regret over wall-clock time on the {\it electricity} dataset for 4 workers (left) and the NASBench201 CIFAR-10 dataset for 8 workers (right). Qualitatively similar results for other datasets can be found in \ifappendix \appref{results}.\else the supplemental material.\fi

\begin{figure*}[t]
\begin{center}
 \includegraphics[width=.49\textwidth]{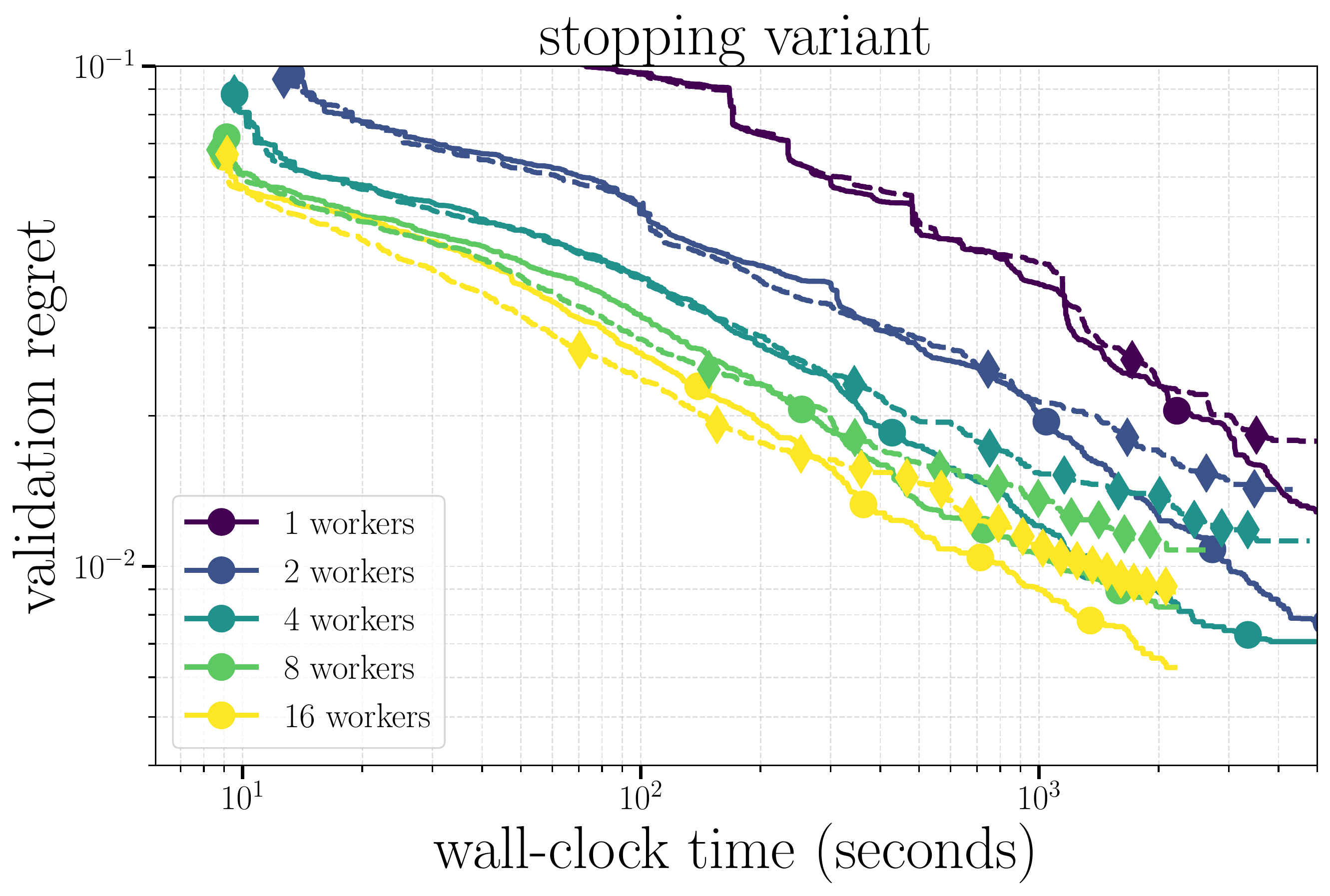}
 \includegraphics[width=.49\textwidth]{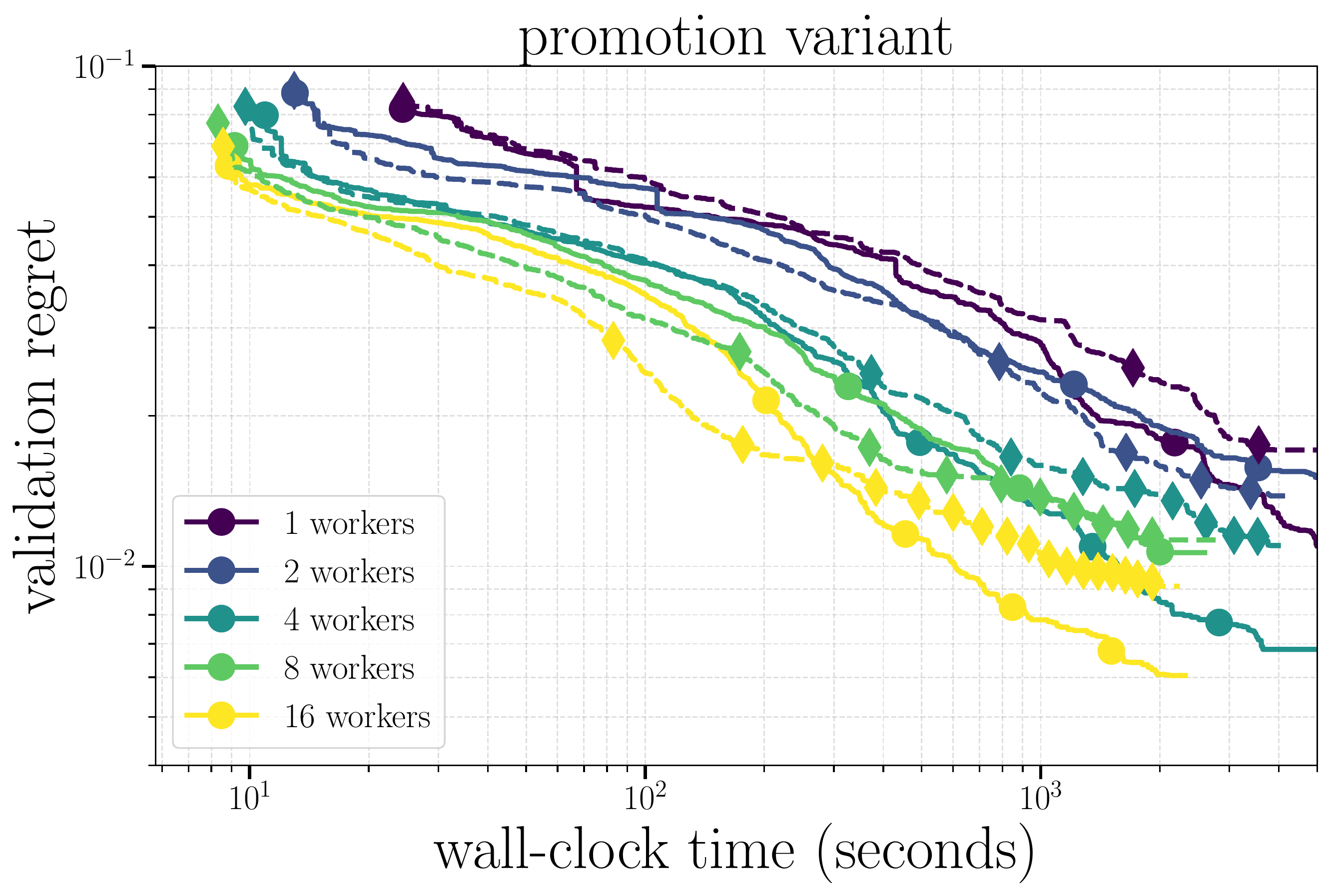}

 \caption{Comparison of stopping (left) and promotion (right) variant for A-HB (dashed lines) and A-BOHB (solid lines) with a variable number of workers on the MLP benchmark. Model-based approaches consistently outperform random-based ones, even as the number of workers increases. }
	\label{fig:async_methods_scaling}
\end{center}
\end{figure*}

{\bf Model-based versus random-based}. First, we focus on differences between our model-based A-BOHB and A-HB, for both scheduling variants. Initially, the model-based variants perform similar to their random-based counterparts. After some time, they rapidly approach the global optimum, while this would take a lot longer for methods relying on random sampling. In line with typical improvements of BO over random search, a well-fitted model focusses the search to a well-performing part of the space that is less likely to be hit by random sampling. As seen in \figref{async_methods_scaling}, the substantial speed-ups of BOHB over HB are observed consistently with 1, 2, 4, 8, and 16 workers on the MLP benchmark. In fact, in most cases A-BOHB converges to a high-accuracy solution in about the same time as A-HB needs, with twice the number of workers: {\em using a joint surrogate model can save half the resources compared to random search}.

{\bf Stopping versus promotion (ASHA) scheduling}. Next, we focus on the two variants of asynchronous scheduling (stopping and promotion), for both A-HB and A-BOHB. On the cheaper MLP benchmark, both scheduling variants perform on-par (\figref{async_methods}, left), while on the more expensive NAS benchmark, the stopping rule achieves a good performance faster than the promotion variant (\figref{async_methods}, right). Recall from \secref{schedule} that for the stopping rule, the initial $\eta$ configurations are not stopped, while the promotion variant pauses all initial configurations at the lowest rung level of their bracket. At least in our expensive experiments, this more careful approach of promoting configurations leads to a worse any-time performance. It is important to note that {\em synchronous} HB and BOHB share this conservative property, along with idle time at synchronization points. At least in situations where initial random configurations, or even well-chosen defaults, provide valuable information at all resource levels, the simpler stopping rule can have an edge. Importantly, in none of our experiments did we witness the initial optimistic scheduling of the stopping rule resulting in worse performance than ASHA at larger wall-clock times.

\subsection{Comparison to State-of-the-Art Methods}

We now compare our novel A-BOHB (stopping) method against a range of state-of-the-art algorithms: asynchronous HB (ASHA) as proposed by Li et al.~\citep{li-arxiv18}, asynchronous Bayesian optimization~\citep{Snoek:12} (A-BO), asynchronous random search~\citep{Bergstra:11} (A-RS), synchronous BOHB~\citep{falkner-icml18} and synchronous HB~\citep{li-iclr17}. Recall that the differences between synchronous and asynchronous BOHB go much beyond scheduling (see \secref{related_work}).

\begin{figure*}[t]
\begin{center}
 \includegraphics[width=.32\linewidth]{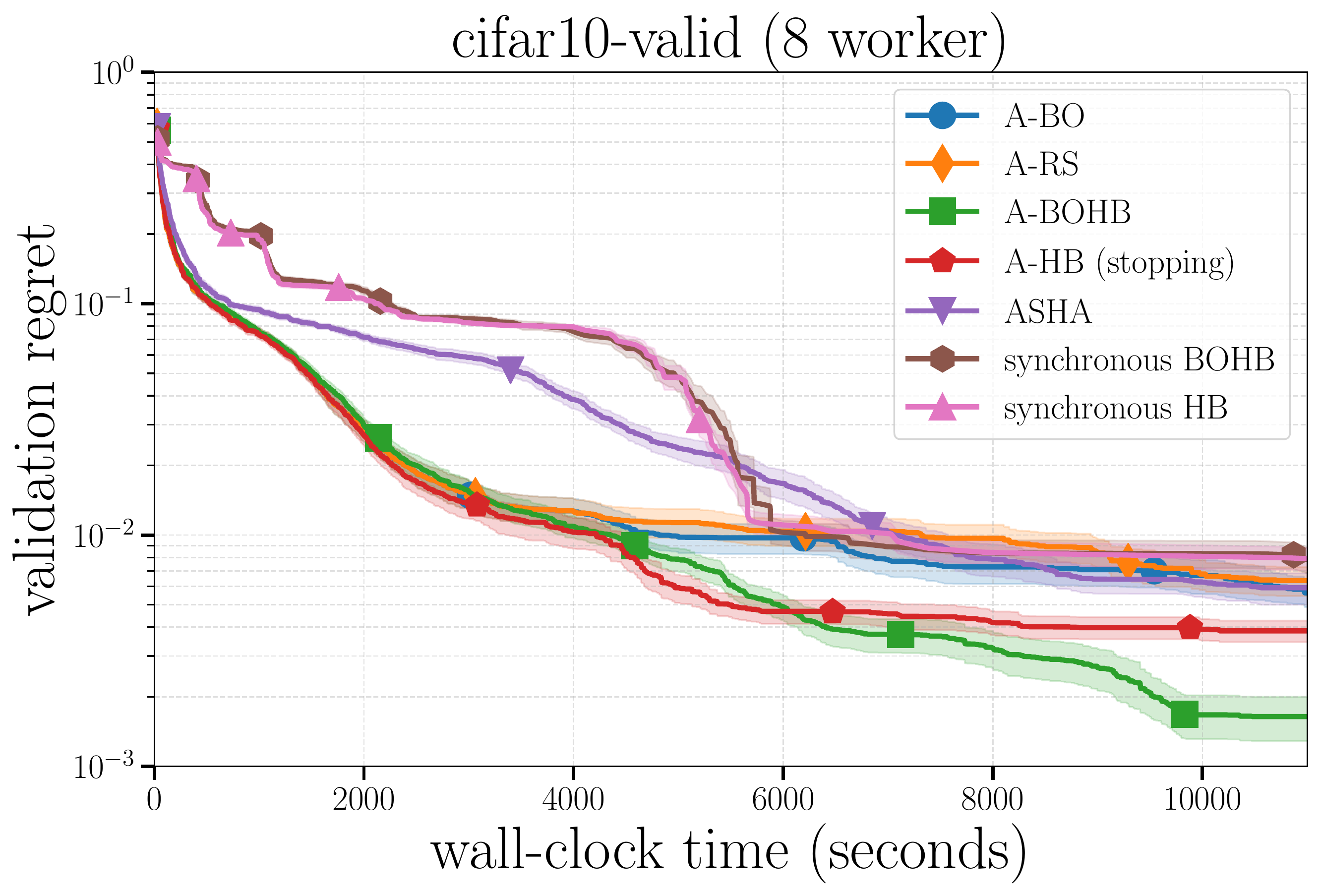}
 \includegraphics[width=.32\linewidth]{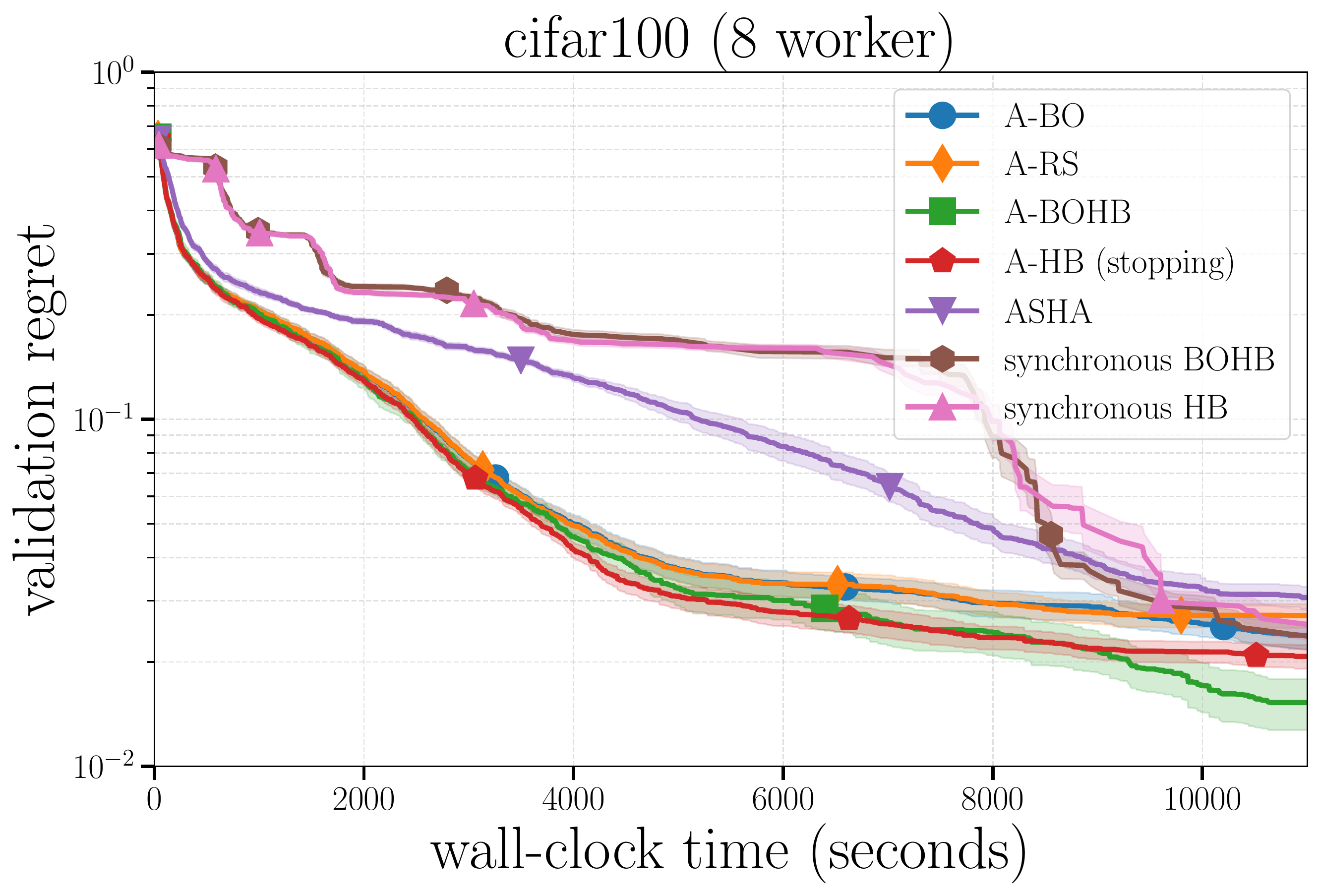}
 \includegraphics[width=.32\linewidth]{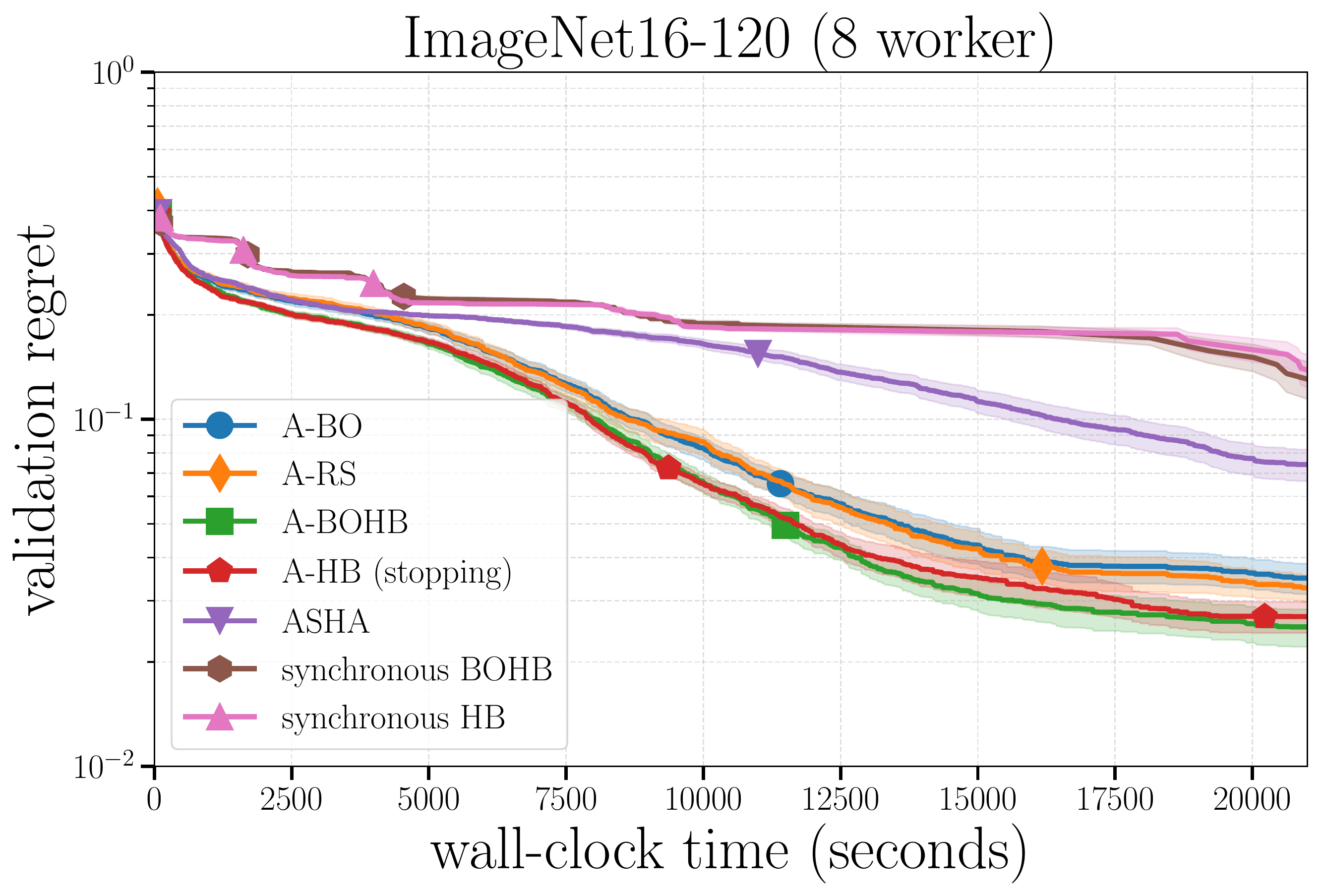}\\ 
	\caption{Comparison of state-of-the-art distributed HNAS methods on different NASBench201 datasets, run with 8 workers. The synchronous methods clearly require more time to reach good accuracies than the asynchronous ones. ASHA is initially slowed down by its more conservative promotion scheme. At higher wall-clock times, our A-BOHB clearly outperforms all other methods.}
\label{fig:nasbench}
\end{center}
\end{figure*}

{\bf NASBench201 datasets.} \figref{nasbench} provides results for the three NASBench201 datasets. ASHA performs better than synchronous HB and BOHB, which also has been observed by Li et al.~\citep{li-arxiv18}, but still requires substantially more time to converge to a sensible performance than BO and RS. Note that metrics are reported after each epoch for any method (BO and RS just cannot stop or pause a configuration), so they all can exhibit good any-time performance. We already observed an edge of stopping over promotion scheduling, yet the effect here is stronger. An explanation could be the observation by Dong et al. \citep{dong-arXiv20} that ranking correlations between architectures evaluated for less than 100 epochs versus the full 200 epochs is small, and that even randomly sampled configurations in NAS search spaces often result in decent performance~\citep{dong-arXiv20,yang-iclr20}. Not only is ASHA slower at promoting any configuration to higher resources, it also may initially promote suboptimal ones. This property would just as well adversely affect synchronous BOHB and HB. In contrast, A-BO and A-BOHB run initial configurations for 200 epochs, sampling valuable data at all resource levels in between, allowing them to make better choices further down. At larger wall-clock times, A-BOHB clearly outperforms A-BO, which shows that the information gathered at all resource levels leads to efficient early stopping of configurations.

According to results on NASBench201 (CIFAR-10) reported by Dong et al.\citep{dong-arXiv20}, the DARTS-V2 method~\citep{liu-iclr19} achieves a poor validation accuracy of $39.77 \pm 0.00 $ (averaged over 3 runs), the best performing method was regularized evolution \citep{real-icml17a} with validation accuracy of $91.19\pm0.31$ (averaged over 500 runs). Our new proposed method A-BOHB achieves a validation accuracy of $91.3 \pm 0.43$ (averaged over 20 runs). The accuracy $90.82\pm0.53$ reported there for synchronous BOHB matches well with $91.02 \pm 0.0029$ in our experiments. While we provide accuracy as function of wall-clock time for a specified compute budget, the results by Dong et al.\citep{dong-arXiv20} are obtained for vastly different compute and wall-clock time budgets.

\begin{figure*}[t]
\begin{center}
\includegraphics[width=.49\linewidth]{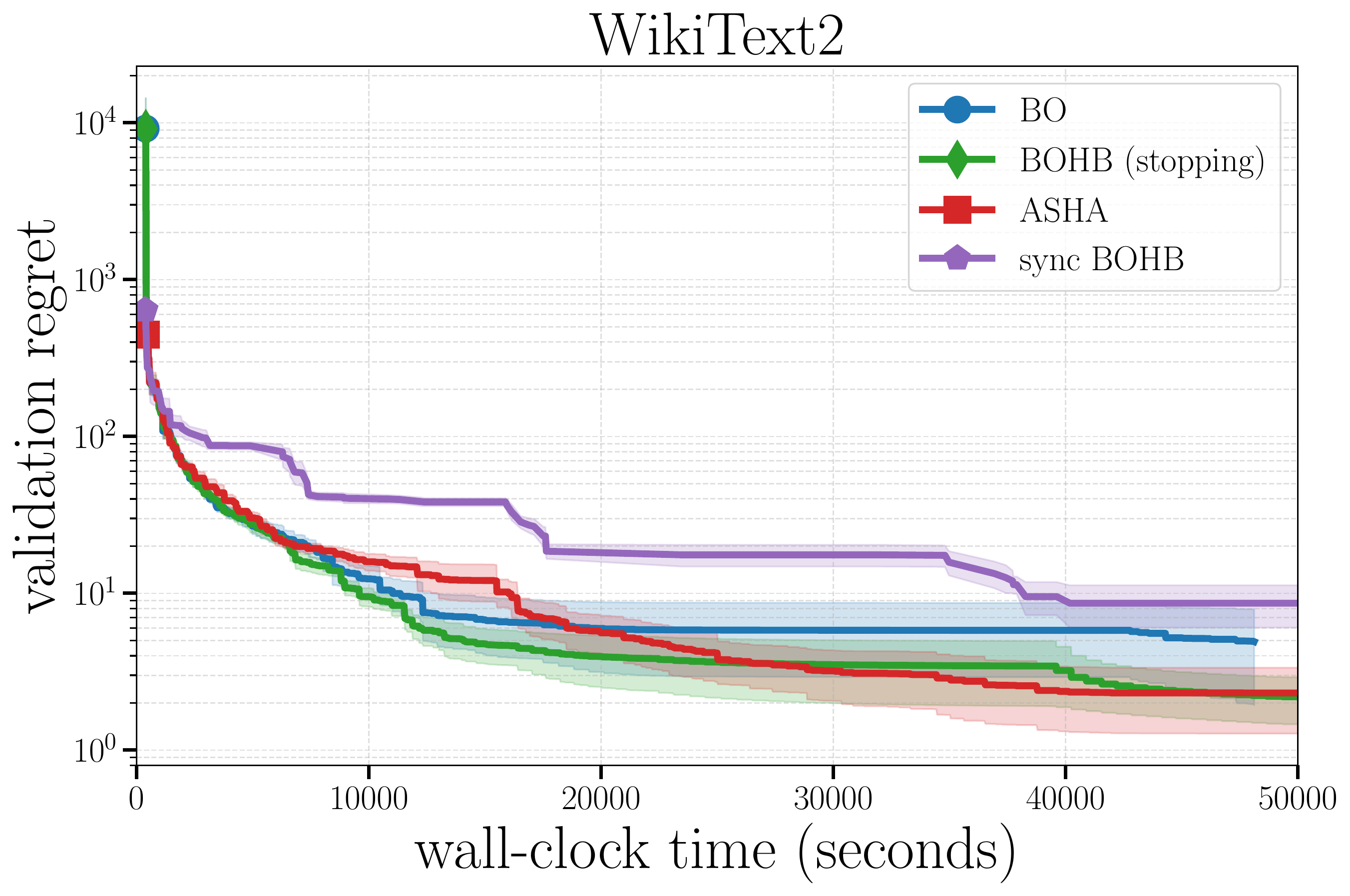}
\includegraphics[width=.49\linewidth]{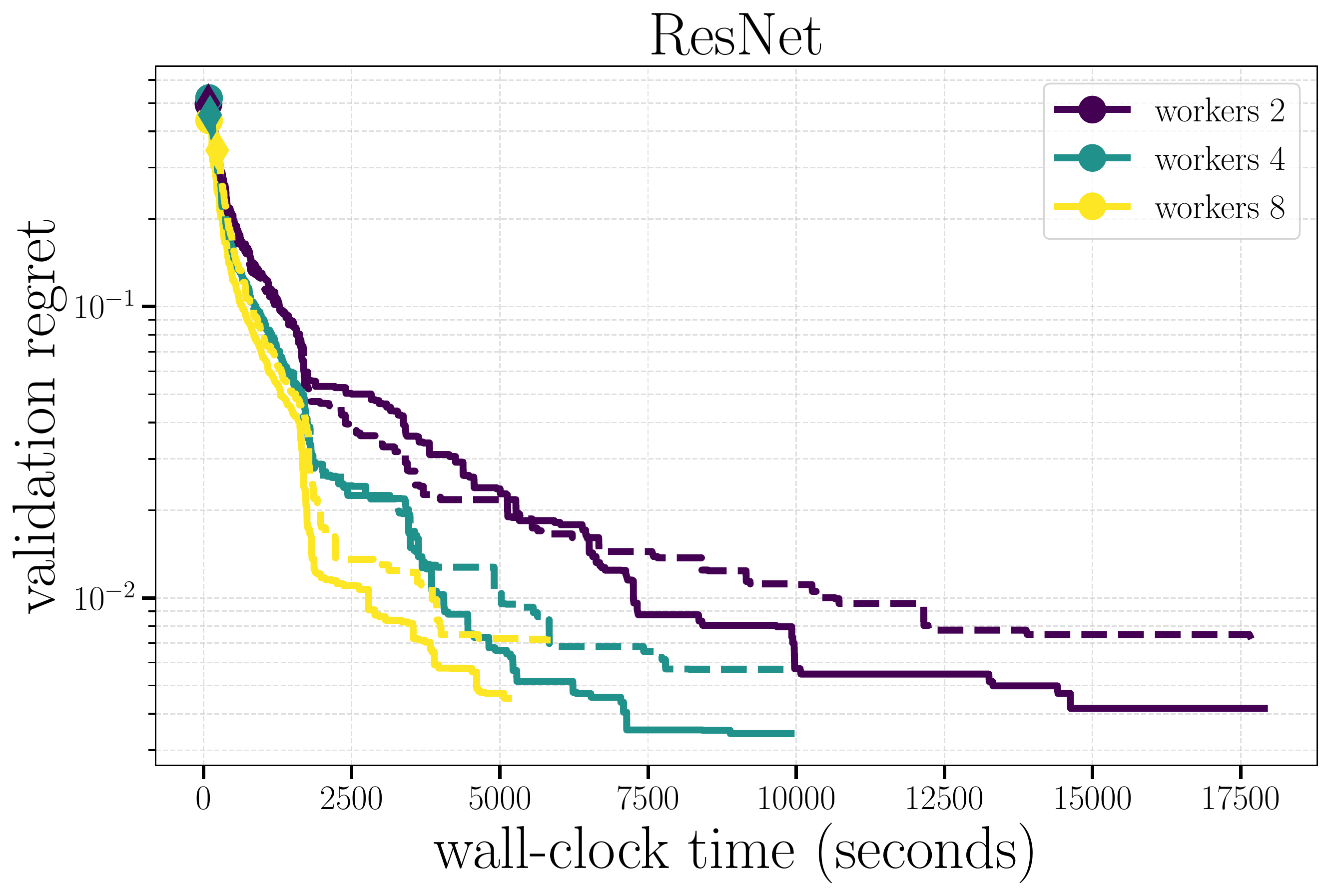}
\caption{\label{fig:wikitext_resnet} Left: Comparison of state-of-the-art distributed HNAS methods on LSTM language modelling (WikiText-2), using 8 workers. Right: Comparison of A-BOHB (bold) and A-HB (dashed), both stopping variant, on tuning ResNet (CIFAR-10).}
\end{center}
\end{figure*}

{\bf LSTM on WikiText-2.} \figref{wikitext_resnet} left shows results for LSTM tuning on the WikiText-2 dataset, using 8 workers. In order to save computations, we limited the comparison to A-BO, A-BOHB (stopping), ASHA, and synchronous BOHB. Training for the full 81 epochs takes roughly 12 hours, and synchronous BOHB is only able to execute a single bracket in the alloted time budget. Due to limited data, it cannot fit a surrogate model at the highest resource level $r_{max}$ (and lacking a joint model, it cannot interpolate from lower rungs either), which may explain its overall worst performance. ASHA performs better on this problem than on the others, and it becomes on-par with A-BOHB after about 25000 secs (circa 7 hours), yet its any-time performance is worse.

{\bf ResNet on CIFAR-10.} In \figref{wikitext_resnet} right, we compare our A-BOHB against A-HB, both with the stopping based variant, 
for 2, 4, and 8 workers respectively. {\em A-BOHB achieves a competitive regret of $10^{-2}$ at roughly the same time as A-HB with twice as many workers} (a regret of $10^{-2}$ means a difference of $1\%$ in validation accuracy).

\section{Conclusions}\label{sec:conclusions}

We present a novel methodology for distributed hyperparameter and architecture search, combining {asynchronous} Hyperband (HB) scheduling with multi-fidelity Bayesian optimization (BO). Instead of sampling new configurations at random, they are chosen based on a joint Gaussian process surrogate model over resource levels, and asynchronous scheduling is accommodated by fantasizing the outcome of pending candidates. While we propose a novel surrogate model, we do not find the kernel choice to be particularly relevant, as long as a probabilistic model interpolates across resources and extends over observed and pending data.

In a range of tuning experiments, spanning MLPs, tabular NAS benchmarks, neural language modelling and image classification, we compare our proposal to prior synchronous and asynchronous HNAS state-of-the-art, focusing on regret as function of wall-clock time (given same resources), which is ultimately most relevant for practitioners. Asynchronous BOHB significantly outperforms both synchronous BOHB and two different variants of asynchronous HB, especially on very expensive benchmarks. The former is slowed down by synchronization overhead, using its parallel resources inefficiently at higher resource levels. We found that asynchronous HB is competitive when given a lot of parallel resources, but degrades with less parallelism. While it finds good solutions, we often find it to plateau before solutions with state-of-the-art performance are found. The asynchronously distributed HNAS method we implemented for this work will be released as open source.

In the future, we will research more principled ways of non-myopic decision-making in the context of asynchronous parallel scheduling, as well as ways to make model-based decisions on scheduling that are more sophisticated than successive halving. We will further investigate advanced bracket sampling strategies and rung specifications.

\clearpage
\newpage

\newpage
\section*{Broader Impact}

Our work encourages the widespread adoption and democratization of AI and 
machine learning.  Without a high degree of domain-specific knowledge and experience, one typically has to resort to laborious manual tuning or brute-force search which, in addition to consuming time and compute resources, can dramatically increase carbon footprint. By offering an efficient and principled approach to automated hyperparameter tuning, we make it accessible for a broader technical audience to effectively apply and deploy off-the-shelf machine learning methods, while utilizing resources more efficiently.

There has been recent massive interest and investment in distributed stochastic optimization applied to neural architecture search (NAS), or more general the automated tuning of expensive deep learning models. While Bayesian optimization (BO) is standard for ML hyperparameter tuning, it has seen surpringly little use in NAS. Most existing open source BO software is limited to single compute instances and mostly sequential search. Asynchronous scheduling in particular, while widely adopted in NAS, is not well served at all. Our work is not only proposing novel methodology for GP-based asynchronous parallel HPO, but also comes with a distributed HPO software system, which will be open sourced along with this publication. Deep learning practitioners with little expertise or interest in BO can use this system in order to automate their workflow, be it tuning hyperparameters or data augmentation, or the search stage of NAS.

\bibliography{amazon,lib,papers,books}

\ifappendix


\clearpage
\newpage

\appendix

\section{Bracket Sampling Distribution}
\label{app:bracket_distribution}

Recall from \secref{schedule} that once a worker is ready to pick up a new task, the corresponding bracket is sampled from a distribution $P(s)$. We use
\begin{equation}
  P(s) \propto \frac{K + 1}{K - s  + 1} \eta^{K - s}.
\end{equation}
Note that $P(s)$ is proportional to the number of configs started in bracket $s$ in synchronous Hyperband~\cite{li-iclr17}. Focussing on the stopping variant, if the rung size distributions were the same as in synchronous HB, this choice would spend the same amount of aggregate resources in each bracket on average. See \figref{bracket_distribution} for an example.
\begin{figure}[h]
  \centering
  \begin{tikzpicture}
  \begin{axis}[
      axis lines = left,
      xlabel = $s$,
      ylabel = {$Z \cdot P(s)$},
  ]
  \addplot [
      domain=0:5, 
      samples=100, 
      color=blue,
      ]
      {(5 + 1)/(5 - x + 1) * 3^(5 - x)};
  \addlegendentry{$K=5, \eta=3$} 
  \end{axis}
  \end{tikzpicture}
  \caption{Bracket distribution (unnormalized) for $K=5, \eta=3$.}
  \label{fig:bracket_distribution}
\end{figure}
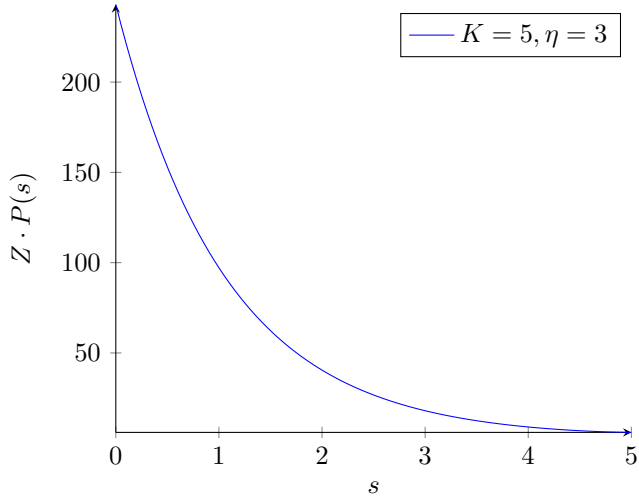

\section{Further Details about the Model}
\label{app:model}

Here, we provide some additional details in order to support the main text.

\subsection{Posterior Representation. Fantasizing}
\label{app:posterior}

We employ a standard representation of the Gaussian process posterior \cite{Rasmussen:06}. Say that $k(\mbx,\mbx')$ is the kernel function, $\mu(\mbx)$ the mean function, and the dataset $\mathcal{D}$ is given by $\mbX\in\bbR^{n\times d}$, $\mby\in\bbR^n$. For notational simplicity, $\mbx$ contains the resource attribute $r$ here. The representation maintains $\mbL$ and $\mbp$, where
\[
   \mbL\mbL^T = \mbK + \sigma^2\mbI_n,\quad \mbp = \mbL^{-1}(\mby -
   \mu(\mbX)).
\]
Here, $\mbK = [k(\mbx_i,\mbx_j)]\in\bbR^{n\times n}$ is the kernel matrix over the training inputs $\mbX$, $\sigma^2$ is the noise variance, and $\mbL$ is the Cholesky factor (lower triangular, positive diagonal).

Recall that we use {\em fantasizing}~\cite{Snoek:12} to deal with pending feedback. For $M$ fantasy samples, we now have a target {\em matrix} $\mbY\in\bbR^{n\times M}$. For any observed configuration $\mbx_i$, the corresponding row in $\mbY$ is $y_i \boldsymbol{1}_M^T$. Now, $\mbL$ remains the same, and $\mbP = \mbL^{-1}(\mbY - \mu(\mbX)\boldsymbol{1}_M^T)$. Importantly, the dominating computation of $\mbL$ is independent of the number $M$ of fantasy samples. Predictions on test points result in $M$ different posterior means, but common posterior variances.

In practice, the surrogate model (kernel and mean function) comes with hyperparameters ({\it e.g.}, noise variance $\sigma^2$, length scales and variance of kernel $k_{\mathcal{X}}$, $\alpha, \beta, \gamma, \delta$), which have to be adjusted. We currently use empirical Bayesian optimization \cite{Rasmussen:06} to this end. All in all, recomputing the posterior representation involves (a) optimizing the hyperparameters on the observed dataset only, and (b) drawing fantasy targets, then computing the representation on the extended dataset, including pending configurations. Here, (a) is substantially more expensive than (b), as it involves computing the representation many times during the optimization.

In our implementation, we delay computations of the posterior representation until the next configuration has to be chosen. If running tasks report metric values, they are inserted into the dataset, replacing pending configurations entered there before. Our implementation offers a number of strategies to skip the expensive part (a) of the computation. Here, (a) is not skipped initially, until the dataset size passes a threshold. After that, we can skip (a) for all but every $k$-th update. Another strategy is to skip (a) as long as the number of {\em full resource} datapoints $(\mbx_i, r_i = r_{\text{max}})$ does not grow. This ensures that we do not spend more time on surrogate model updates in the multi-fidelity case than in standard BO with full resource evaluations only.

\subsection{Exponential Decay Surrogate Model}
\label{app:surrogate}

The model proposed by Swersky et al.~\citep{Swersky:14} is based on the assumption $f(\mbx, r) = f(\mbx) + e^{-\lambda r}$, where $\lambda\sim p(\lambda)$ is drawn from a Gamma prior, and the asymptote $f(\mbx)$ ({\it i.e.}, $f(\mbx, r)$ as $r\to \infty$) from a GP prior. They end up using a zero-mean GP prior with kernel function $k((\mbx, r), (\mbx', r')) = k_{\mathcal{X}}(\mbx, \mbx') + k_{\mathcal{R}}(r, r')$ and
\begin{align*}
   k_{\mathcal{R}}(r, r') & := \int_0^{\infty} e^{-\lambda r} e^{-\lambda r'} p(\lambda)\,
   d\lambda = \kappa(r + r'), \\
   \kappa(u) & := \frac{\beta^{\alpha}}{(u + \beta)^{\alpha}},\quad \alpha, \beta > 0.
\end{align*}
This proposal has several shortcomings. First, $f(\mbx, r) \to f(\mbx) + 1$ as $r \to 0$, independent of what metric is used. Second, the random function $e^{-\lambda r}$ assumption implies a non-zero mean function, {and a kernel taking this into account. A better assumption would be $f(\mbx, r) = f(\mbx) + \gamma e^{-\lambda r}$, $\gamma>0$, which implies a mean function $\mu(\mbx, r) = \mu_{\mathcal{X}}(\mbx) + \gamma\kappa(r)$ and a covariance function
\begin{align}
   k((\mbx, r), (\mbx', r')) & = k_{\mathcal{X}}(\mbx, \mbx') + \gamma^2
   \tilde{k}_{\mathcal{R}}(r, r'), \label{eq:additive} \\
   \tilde{k}_{\mathcal{R}}(r, r') & := \kappa(r + r') - \kappa(r) \kappa(r'). \nonumber
\end{align}
One advantage of this additive model structure is that an implied conditional independence relation can be exploited in order to speed up inference computations \cite{Swersky:14}. However, this model implies $f(\mbx, r) = f(\mbx) + \gamma$ as $r \to 0$: the metric for no training at all depends on $\mbx$ in the same way as the asymptote. A more sensible assumption would be $f(\mbx, r) = \gamma$ as $r \to 0$, since without training, predictions should be random ({\it e.g.}, $\gamma=\frac{1}2$ for binary classification). We present a novel surrogate model which can incorporate this assumption. It is based on
\begin{equation}\label{eq:surrogate-f}
   f(\mbx, r) = \gamma e^{-\lambda r} + f(\mbx) \left( 1 - \delta e^{-\lambda r}
   \right),\quad \delta\in [0, 1].
\end{equation}
Some algebra (provided below) gives
\begin{align*}
  \mu(\mbx, r) & = \gamma\kappa(r) + \mu_{\mathcal{X}}(\mbx) (1 - \delta\kappa(r)), \\
  \begin{split}
     k((\mbx, r), (\mbx', r')) 
     & = (\gamma - \delta\mu_{\mathcal{X}}(\mbx))
     \tilde{k}_{\mathcal{R}} (r, r') (\gamma - \delta\mu_{\mathcal{X}}(\mbx')) \\
     & \quad + k_{\mathcal{X}}(\mbx, \mbx') [ 1 - \delta(\kappa(r) + \kappa(r') \\
     & \quad - \delta\kappa(r + r')) ].    
  \end{split}
\end{align*}
For $\delta=0$, we recover the additive model of \eqref{additive}, while for $\delta=1$, we have that $f(\mbx, r) = \gamma$ as $r \to 0$, independent of $\mbx$. This model comes with hyperparameters $\alpha, \beta, \gamma > 0$ and $\delta\in[0, 1]$.
We provide an empirical analysis of this model in \ifappendix \appref{surrogate_models}.\else the supplemental material.\fi

\subsubsection*{Derivation of Exponential Decay Surrogate Model}

Recall that our novel surrogate model $k((\mbx, r), (\mbx',r'))$ and $\mu(\mbx, r)$ is based on the random function assumption \eqp{surrogate-f}. Since $\mathbb{E}[e^{-\lambda r}] = \kappa(r)$, it is clear that $\mathbb{E}[f(\mbx, r)] = \gamma\kappa(r) + \mu_{\mathcal{X}}(\mbx)(1 - \delta\kappa(r)) = \mu_{\mathcal{X}}(\mbx) + (\gamma - \delta \mu_{\mathcal{X}}(\mbx)) \kappa(r)$. Moreover,
\[
\begin{split}
   & k((\mbx, r), (\mbx', r')) = \mathbb{E}\left[ f(\mbx, r) f(\mbx', r') \right] \\
   & =\, (\gamma - \delta\mu_{\mathcal{X}}(\mbx)) \tilde{k}_{\mathcal{R}}(r, r') (\gamma - \delta\mu_{\mathcal{X}}(\mbx')) + k_{\mathcal{X}}(\mbx, \mbx') \\
   & \cdot\, \mathbb{E}\left[ \left( 1 -\delta e^{-\lambda r} \right) \left( 1 -\delta e^{-\lambda r'} \right) \right] = (\gamma - \delta\mu_{\mathcal{X}}(\mbx)) \\
   & \cdot\, \tilde{k}_{\mathcal{R}}(r, r') (\gamma - \delta\mu_{\mathcal{X}}(\mbx')) + k_{\mathcal{X}}(\mbx, \mbx') \\
   & \cdot\, \left( 1 - \delta( \kappa(r) + \kappa(r') -
   \delta\kappa(r + r') ) \right), \\
   & \tilde{k}_{\mathcal{R}}(r, r') = \kappa(r + r') - \kappa(r) \kappa(r').
\end{split}
\]

\subsection{Data for Surrogate Model}
\label{app:gp-searcher-data}

What data should be used to fit the surrogate model? Assume that $r$ is the number of training epochs. If not stopped, we observe $f(\mbx, r)$ for $r = r_{\text{min}}\eta^s, r_{\text{min}}\eta^s + 1,\dots$, yet only observations at $r\in B_s$ are taking part in stopping or promotion decisions. In asynchronous BOHB, as detailed in \secref{model}, we condition the GP model only on observations at rung levels $r\in B_s$. Since model computations scale cubically with the data size, this seems sensible to keep costs at bay. Still, our implementation offers two other options. First, as argued by Swersky et al.\citep{Swersky:14}, we may use all observed data. However, this drives up the cost for decision-making. In this case, we register pending evaluations for $(\mbx, r+1)$ once $f(\mbx, r)$ is observed. We also support a third option, which retains observations at $r\in B_s$, as well as the most recent $f(\mbx, r)$ ({\it i.e.}, largest $r$). \figref{data} compares the three different options for asynchronous BOHB (stopping) on the electricity dataset with 4 workers. The experimental setup is the same as described in \secref{experiments} in the main text. Fitting the GP on all the data indeed lowers the overall performance of BOHB. On the other hand, while there is no substantial difference between the first and the third option. Using data at rung levels only is the simplest choice, and we used it for all other experiments.

\begin{figure}[ht]
\begin{center}
\includegraphics[width=\linewidth]{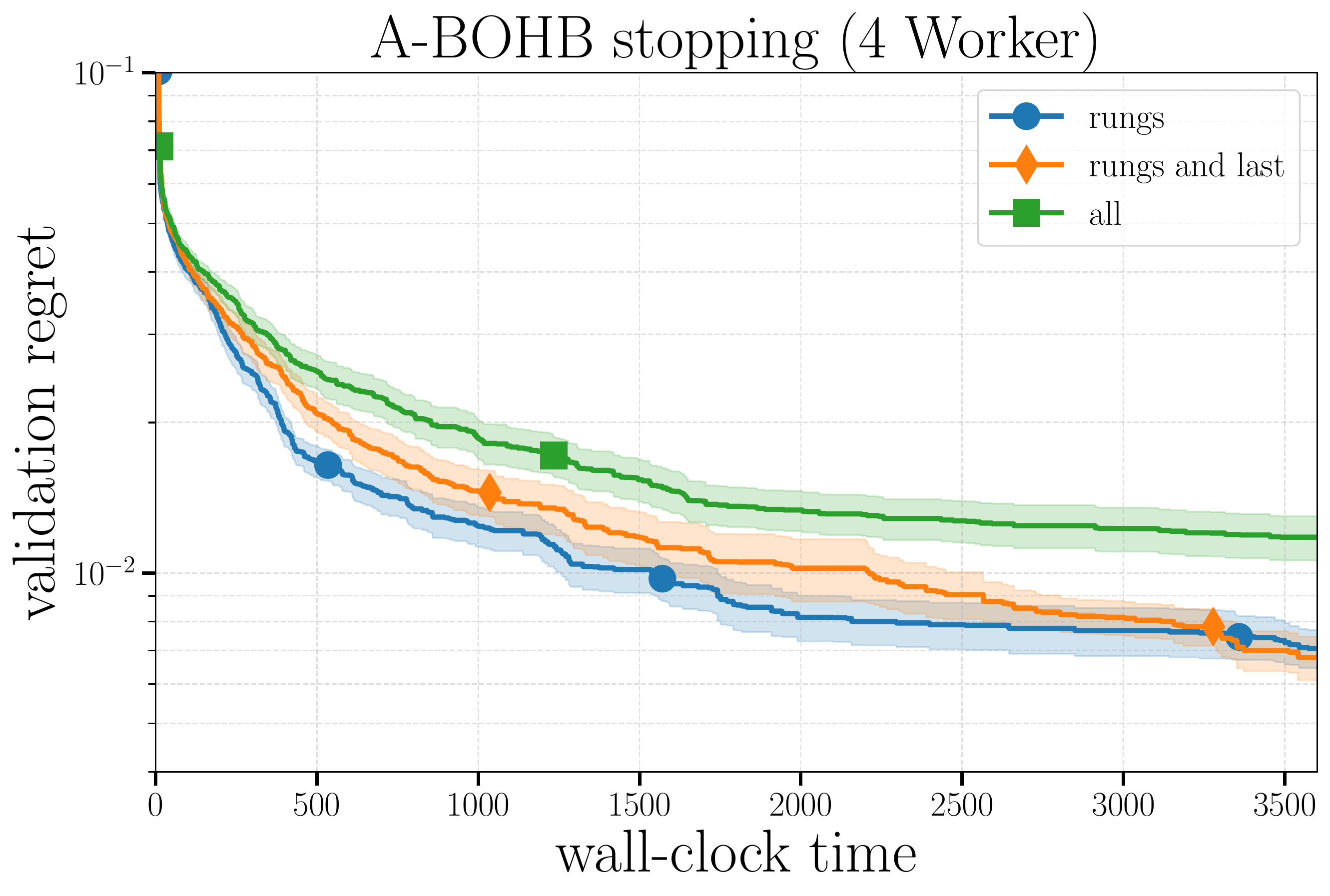}
 
\caption{The figure shows the effect of what data is fit to the surrogate model, where `rungs` means we only used the data at specific rung levels (as done in all experiments in the main paper), for 'rungs and last' we additionally feed the last observed value of each learning curve to the GP, and 'all' means we fit our model on metric data obtained at all resource levels.}
\label{fig:data}
\end{center}
\end{figure}

\section{Configuration Spaces}\label{app:hpranges}

Table~\ref{tab:hp_ranges_mlp} shows the configuration space for the MLP benchmark, Table~\ref{tab:hp_ranges_res_net} for the residual network benchmark, Table~\ref{tab:hp_ranges_nasbench201} for NASBench201 and Table~\ref{tab:hp_ranges_lstm} for the LSTM benchmark.
If we use an exponential notation than we optimized the corresponding hyperparameter on a logarithmic scale. We used a one-hot encoding for all categorical hyperparameters.

\begin{table}[t]
\centering
\caption[Hyperparameters for MLP benchmark]{The hyperparameters and architecture choices for the fully connected networks.}
\begin{tabular}{|c|c|}
\hline
	Hyperparameter & Range \\
	\hline
	learning rate& $[10^{-6}, 1]$  \\
	batch size & $[2^3, 2^7]$ \\
	dropout layer 1 & $[0, .99]$ \\
	dropout layer 2 & $[0, .99]$ \\
	units layer 1 & $[2^4, 2^{10}]$ \\
	units layer 2 & $[2^4, 2^{10}]$ \\
	scale layer 1 & $[10^{-3}, 10]$ \\
	scale layer 2 & $[10^{-3}, 10]$ \\
\hline
\end{tabular}
\label{tab:hp_ranges_mlp}
\end{table}

\begin{table}[t]
\centering
\caption[Hyperparameters for Res-Net benchmark]{The configuration space for the residual neural network benchmark.}
\begin{tabular}{|c|c|}
\hline
	Hyperparameter & Range \\
	\hline
	learning rate& $[1^{-3}, 1^{-1}]$  \\
	batch size & $[8, 256]$ \\
	weight decay & $[10^{-5}, 10^{-3}]$ \\
	momentum & $[0, .99]$ \\
\hline
\end{tabular}
\label{tab:hp_ranges_res_net}
\end{table}

\begin{table}[t]
\centering
\caption[Hyperparameters for LSTM benchmark]{The configuration space for the LSTM benchmark.}
\begin{tabular}{|c|c|}
\hline
	Hyperparameter & Range \\
	\hline
	learning rate& $[1, 50]$  \\
	batch size & $[2^3, 2^7]$ \\
	dropout & $[0, 0.99]$ \\
	gradient clipping & $[0.1, 2]$ \\
	learning rate factor & $[1, 100]$ \\
\hline
\end{tabular}
\label{tab:hp_ranges_lstm}
\end{table}

\begin{table}[t]
\centering
\caption[Hyperparameters for NASBench201]{The configuration space for NASBench201.}
\begin{tabular}{|c|c|}
\hline
	Hyperparameter & Range \\
	\hline
	edge 0 & $\{ none, skip\mhyphen connect, nor\mhyphen conv\mhyphen 1x1, nor\mhyphen conv\mhyphen 3x3, avg\mhyphen pool\mhyphen 3x3 \}$ \\
	edge 1 & $\{ none, skip\mhyphen connect, nor\mhyphen conv\mhyphen 1x1, nor\mhyphen conv\mhyphen 3x3, avg\mhyphen pool\mhyphen 3x3 \}$ \\
	edge 2 & $\{ none, skip\mhyphen connect, nor\mhyphen conv\mhyphen 1x1, nor\mhyphen conv\mhyphen 3x3, avg\mhyphen pool\mhyphen 3x3 \}$ \\
	edge 3 & $\{ none, skip\mhyphen connect, nor\mhyphen conv\mhyphen 1x1, nor\mhyphen conv\mhyphen 3x3, avg\mhyphen pool\mhyphen 3x3 \}$ \\
	edge 4 & $\{ none, skip\mhyphen connect, nor\mhyphen conv\mhyphen 1x1, nor\mhyphen conv\mhyphen 3x3, avg\mhyphen pool\mhyphen 3x3 \}$ \\
	edge 5 & $\{ none, skip\mhyphen connect, nor\mhyphen conv\mhyphen 1x1, nor\mhyphen conv\mhyphen 3x3, avg\mhyphen pool\mhyphen 3x3 \}$ \\

\hline
\end{tabular}
\label{tab:hp_ranges_nasbench201}
\end{table}

\section{Additional Experiments}\label{app:results}

Here we provide additional experiments that extend the experiments from the main paper. 

\subsection{MLP on Classification Datasets}

\figref{mlp_all} shows the results for four different classification datasets from OpenML~\citep{vanschoren-sigkdd13a}: \textit{electricity} (OpenML task ID 336), \textit{helena} (OpenML task ID 168329), \textit{robert} (OpenML task ID 168332) and \textit{dilbert} (OpenML task ID 168909).
As described in the main text, after some initial phase, model-based methods perform better than their random-search based counterparts across all datasets. In contrast to the NASBench201 benchmarks, here the promotion and the stopping variant of A-BOHB perform similarly.

\begin{figure*}[ht]
\begin{center}
 \includegraphics[width=.49\linewidth]{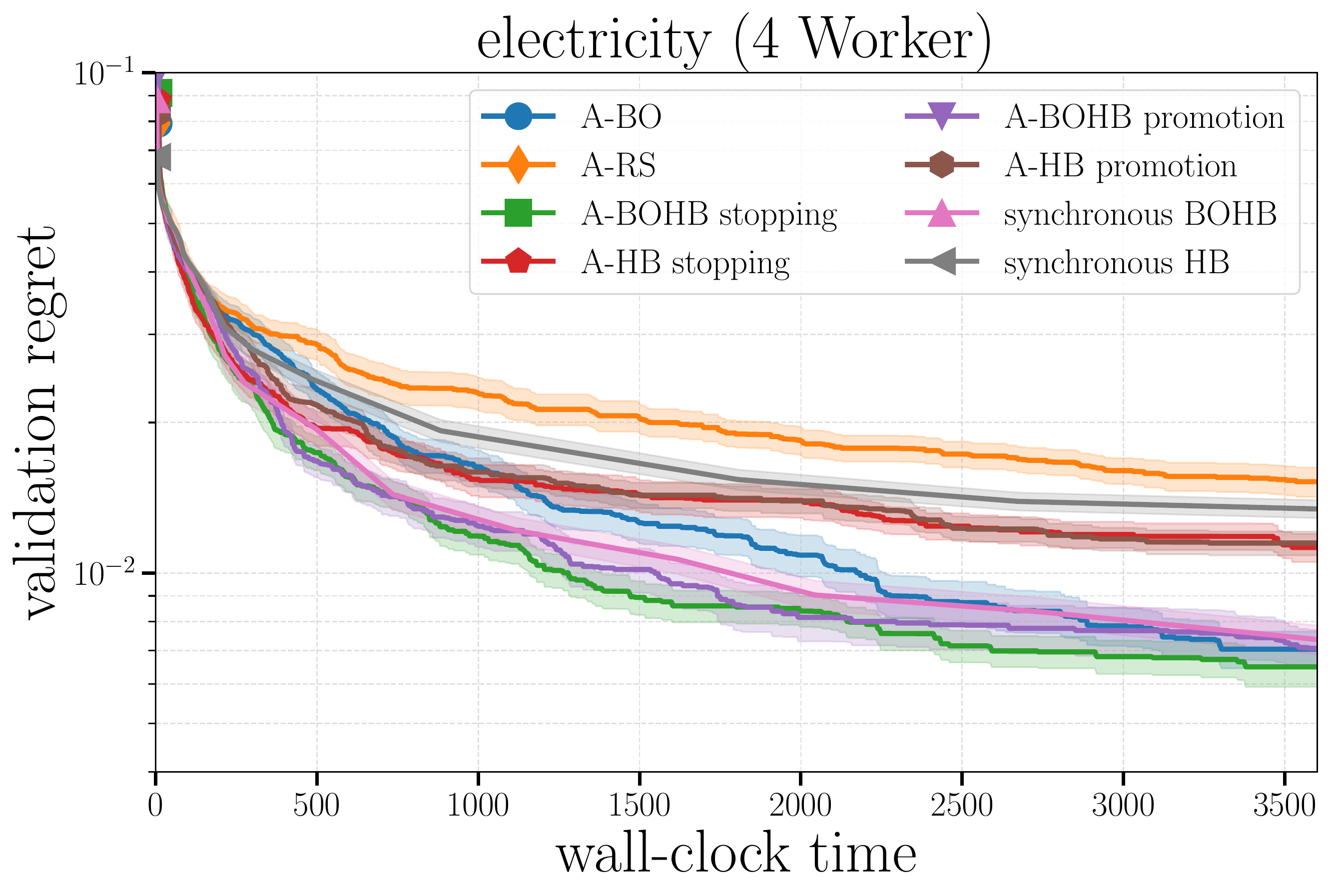}
 \includegraphics[width=.49\linewidth]{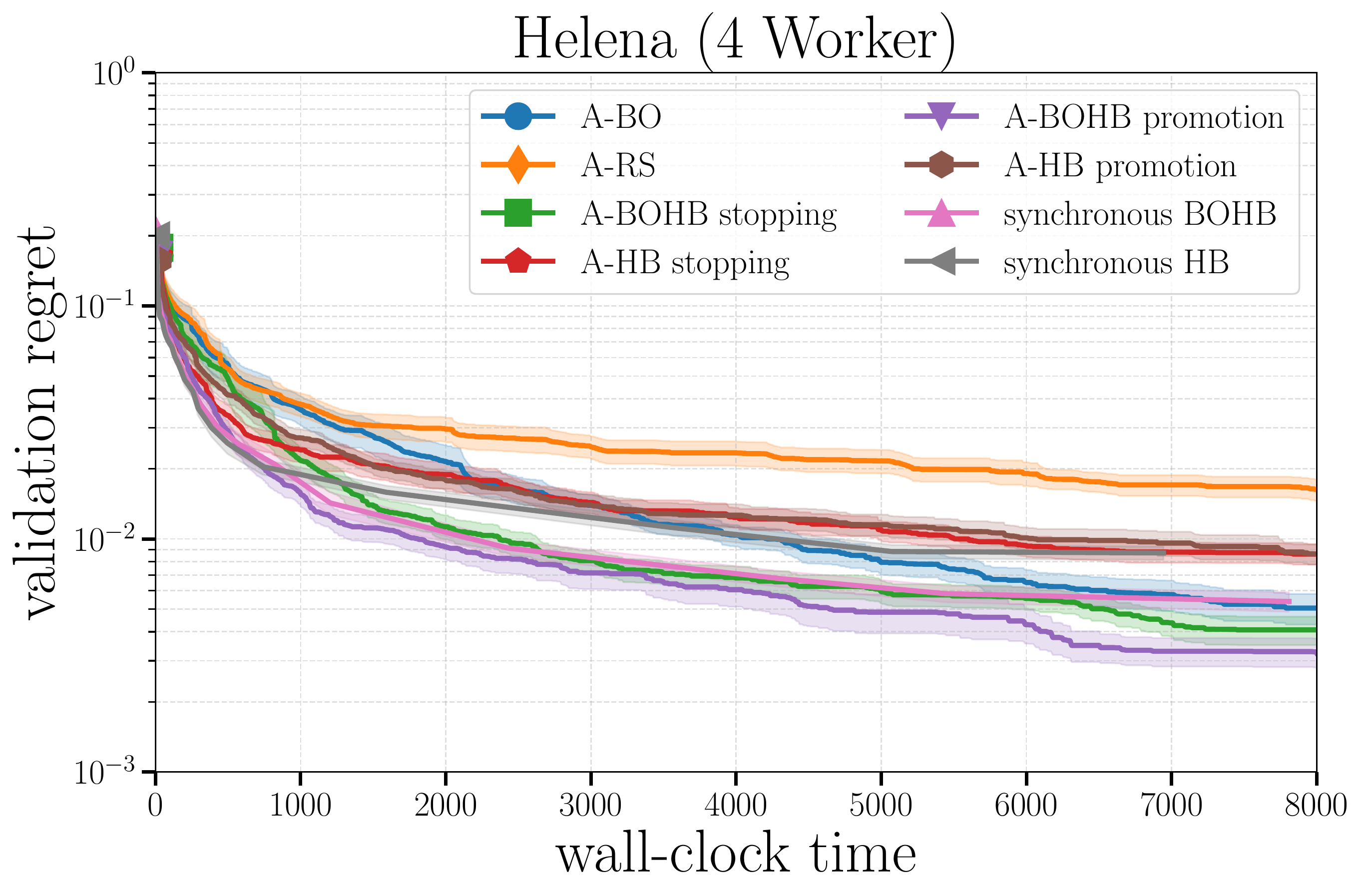}\\
 \includegraphics[width=.49\linewidth]{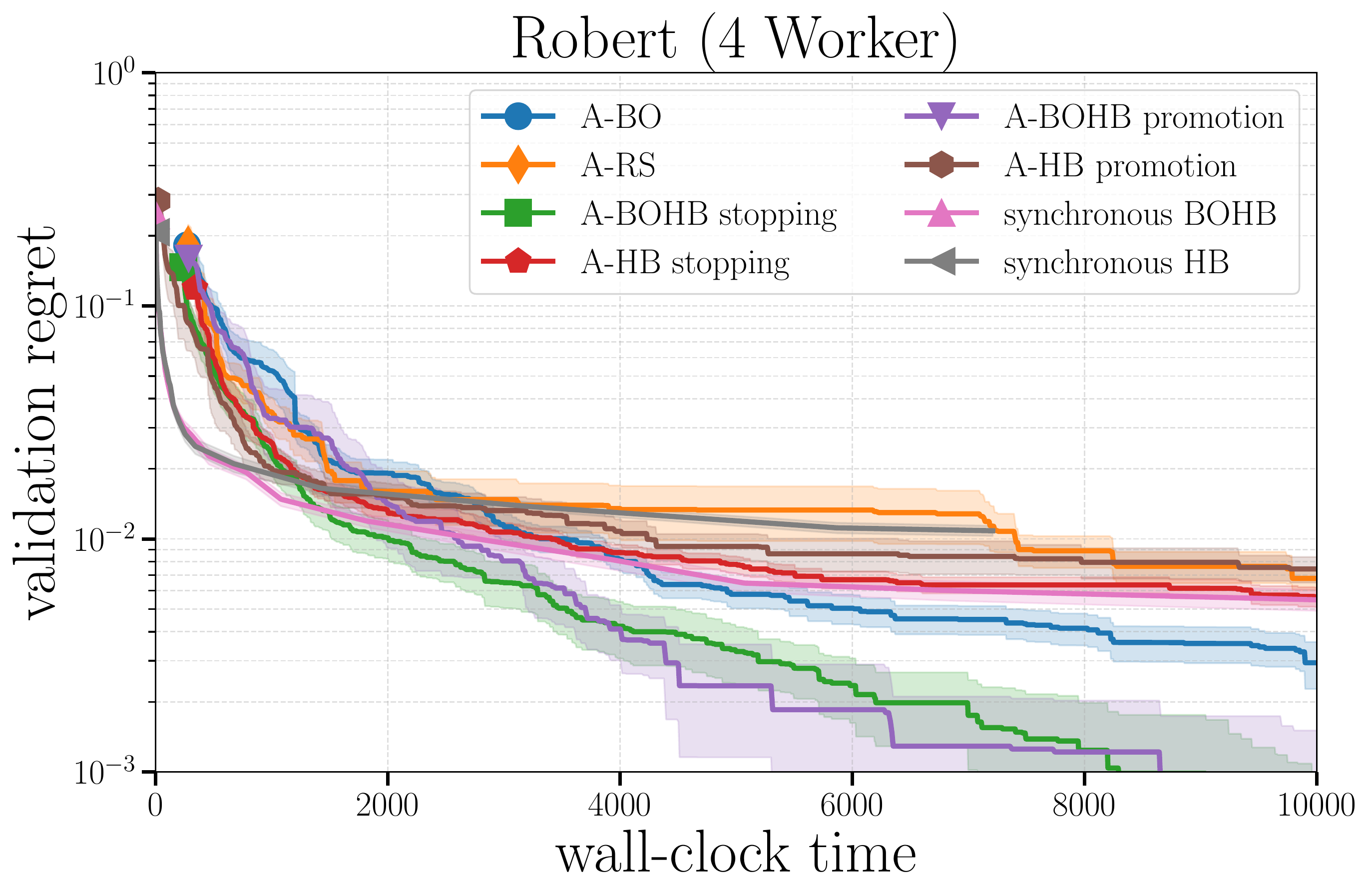}
 \includegraphics[width=.49\linewidth]{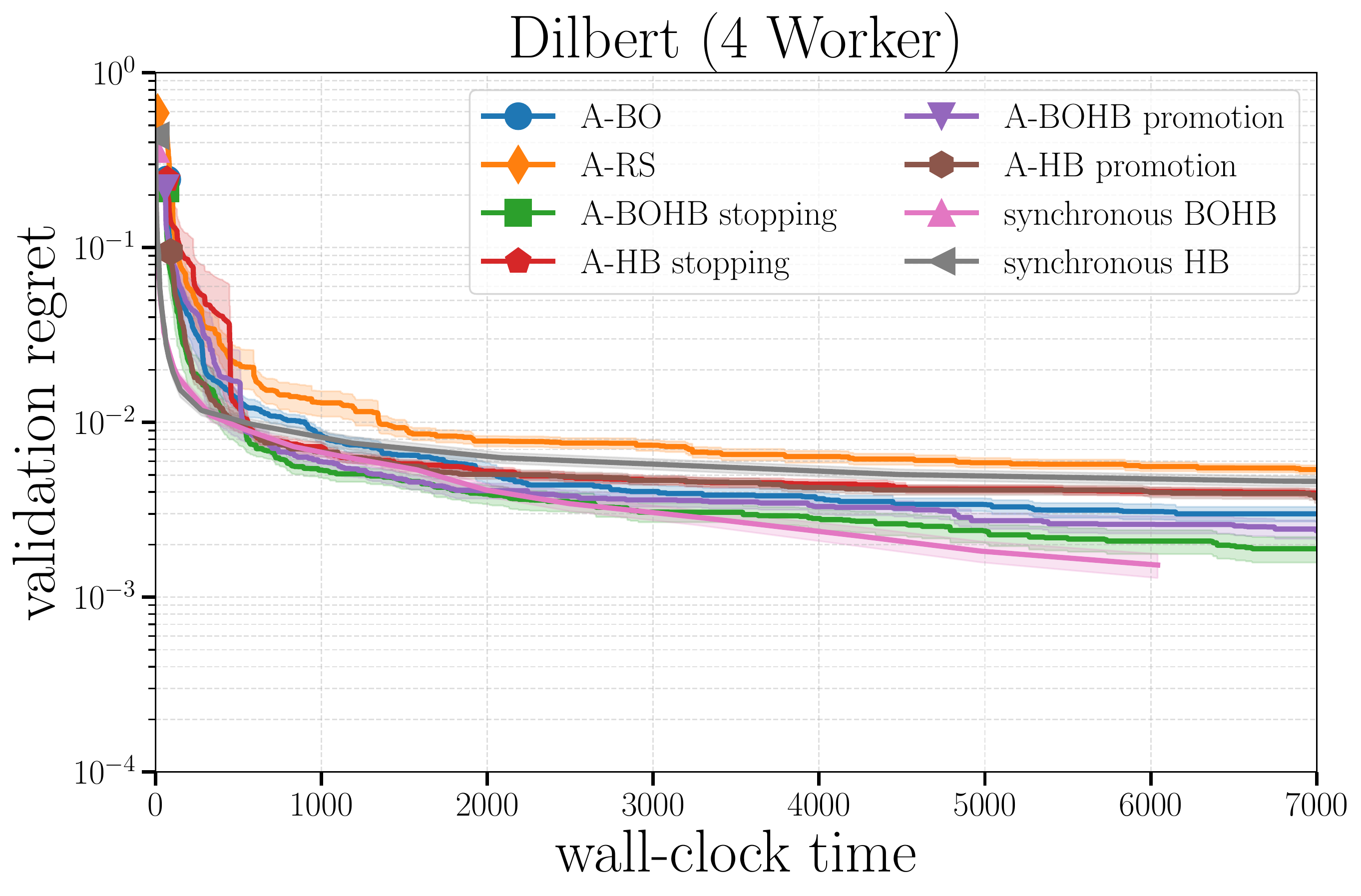}
 \caption{Validation regret of all different HB and BOHB variants on 4 different OpenML classification datasets.}
\label{fig:mlp_all}
\end{center}
\end{figure*}

\subsection{Surrogate Models}
\label{app:surrogate_models}

\begin{figure*}[ht]
\begin{center}
 \includegraphics[width=.32\linewidth]{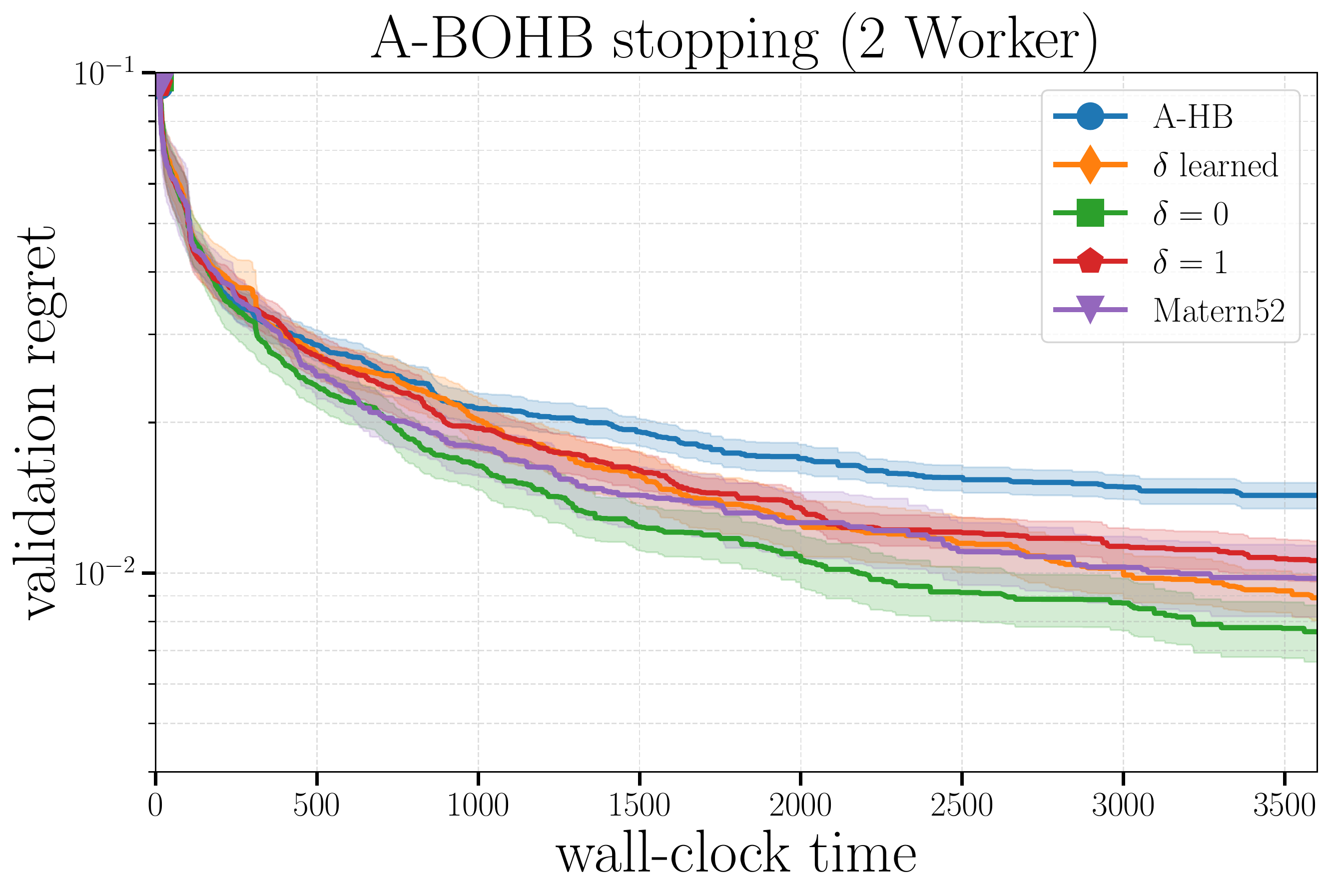}
 \includegraphics[width=.32\linewidth]{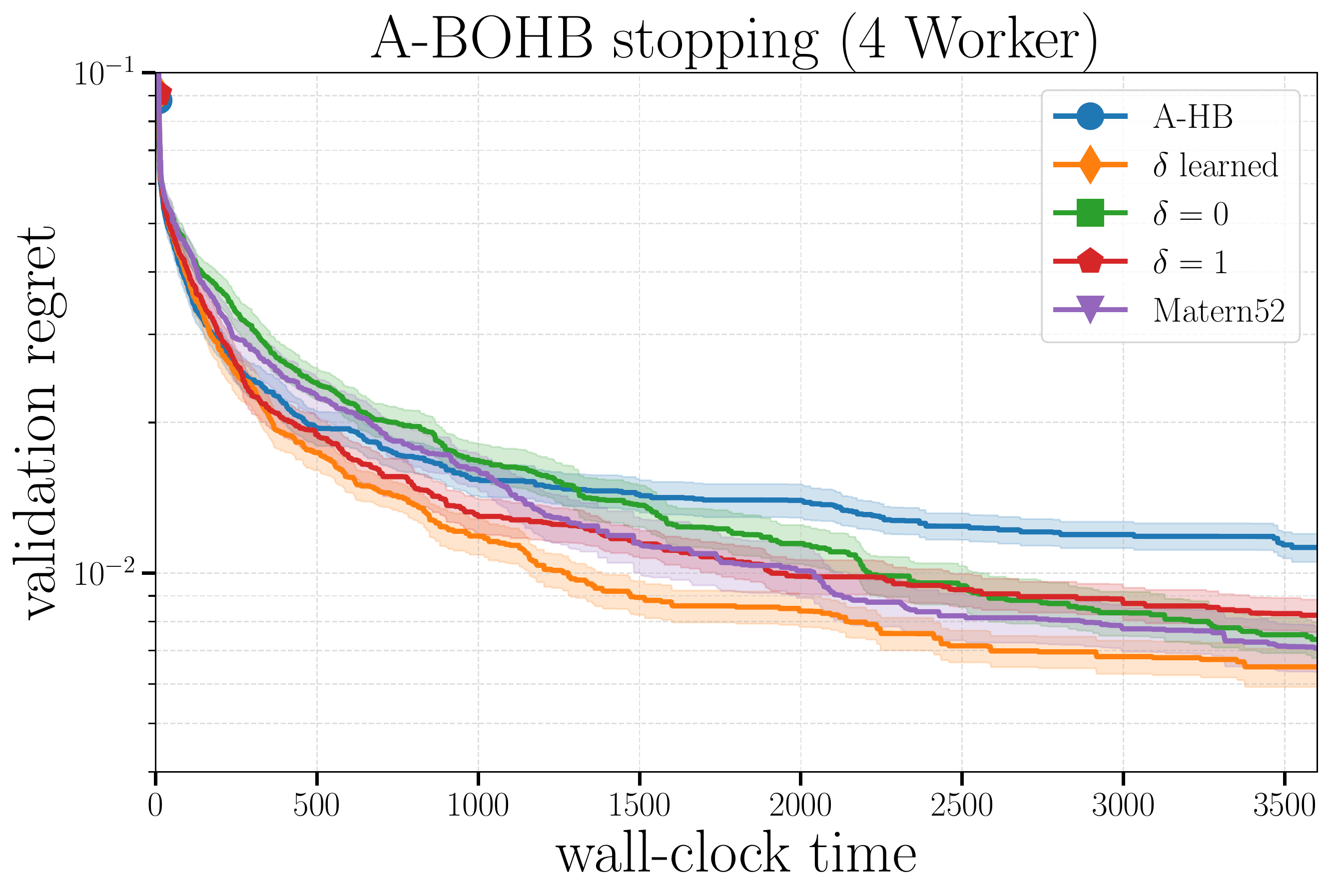}
 \includegraphics[width=.32\linewidth]{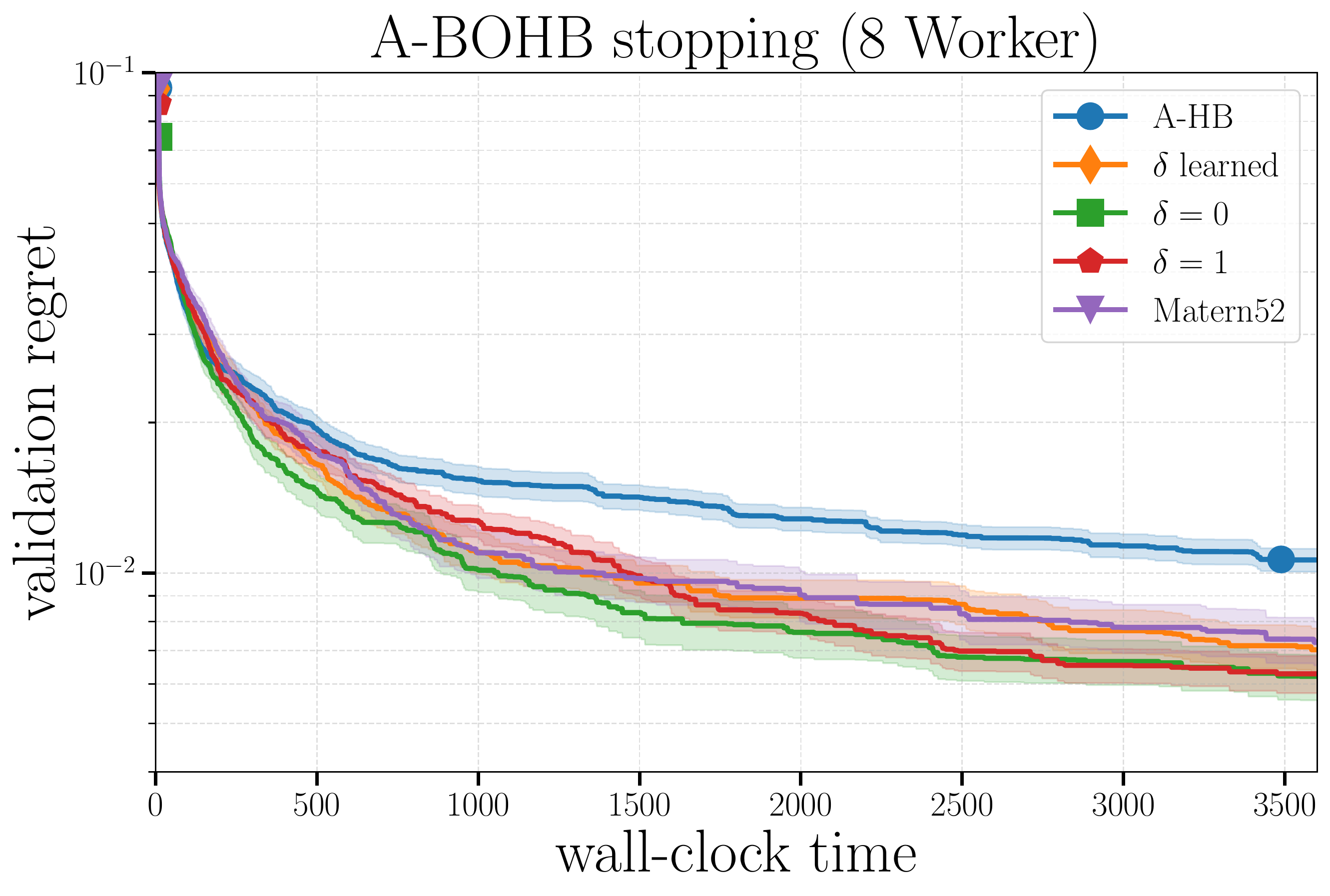}\\
 \includegraphics[width=.32\linewidth]{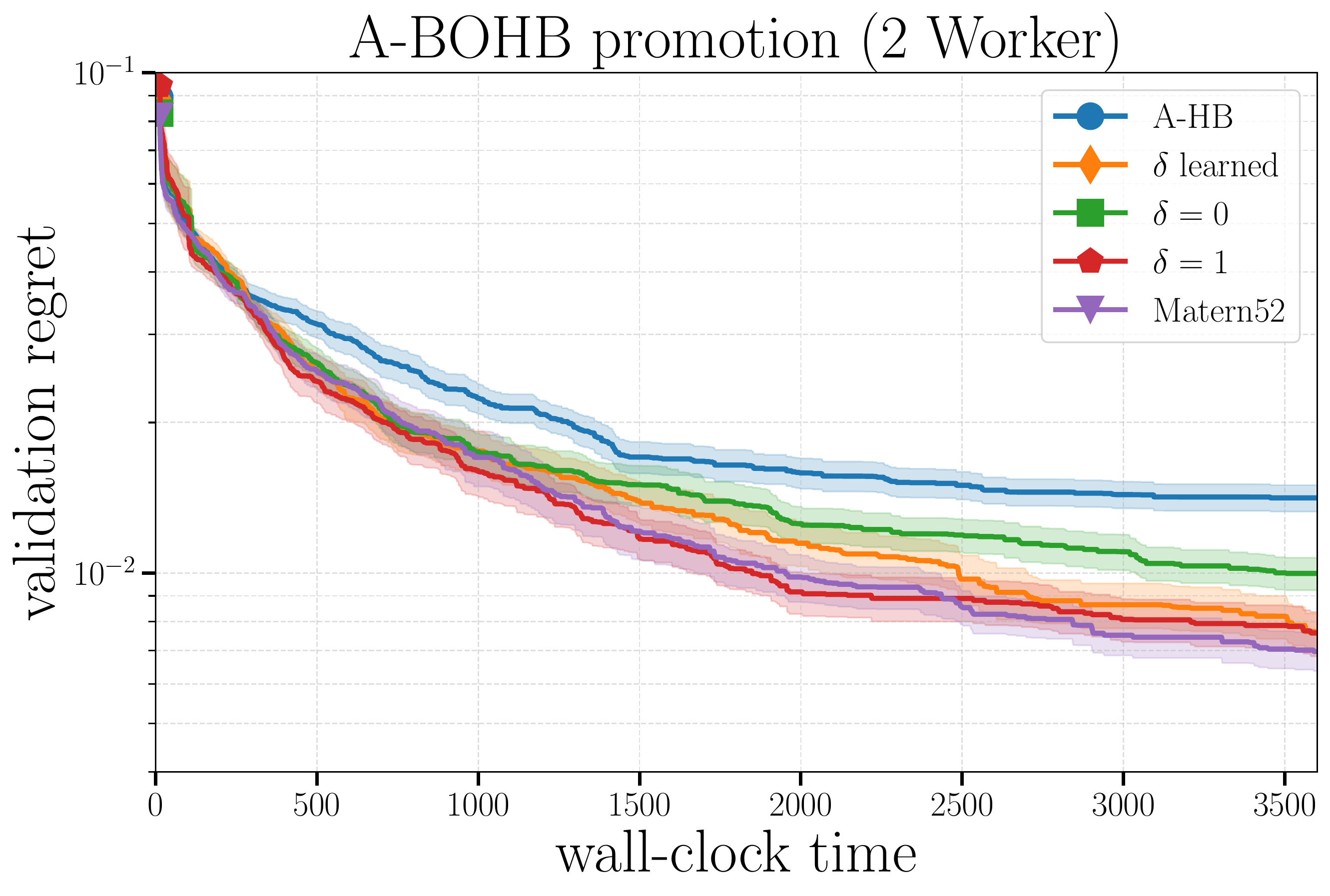}
 \includegraphics[width=.32\linewidth]{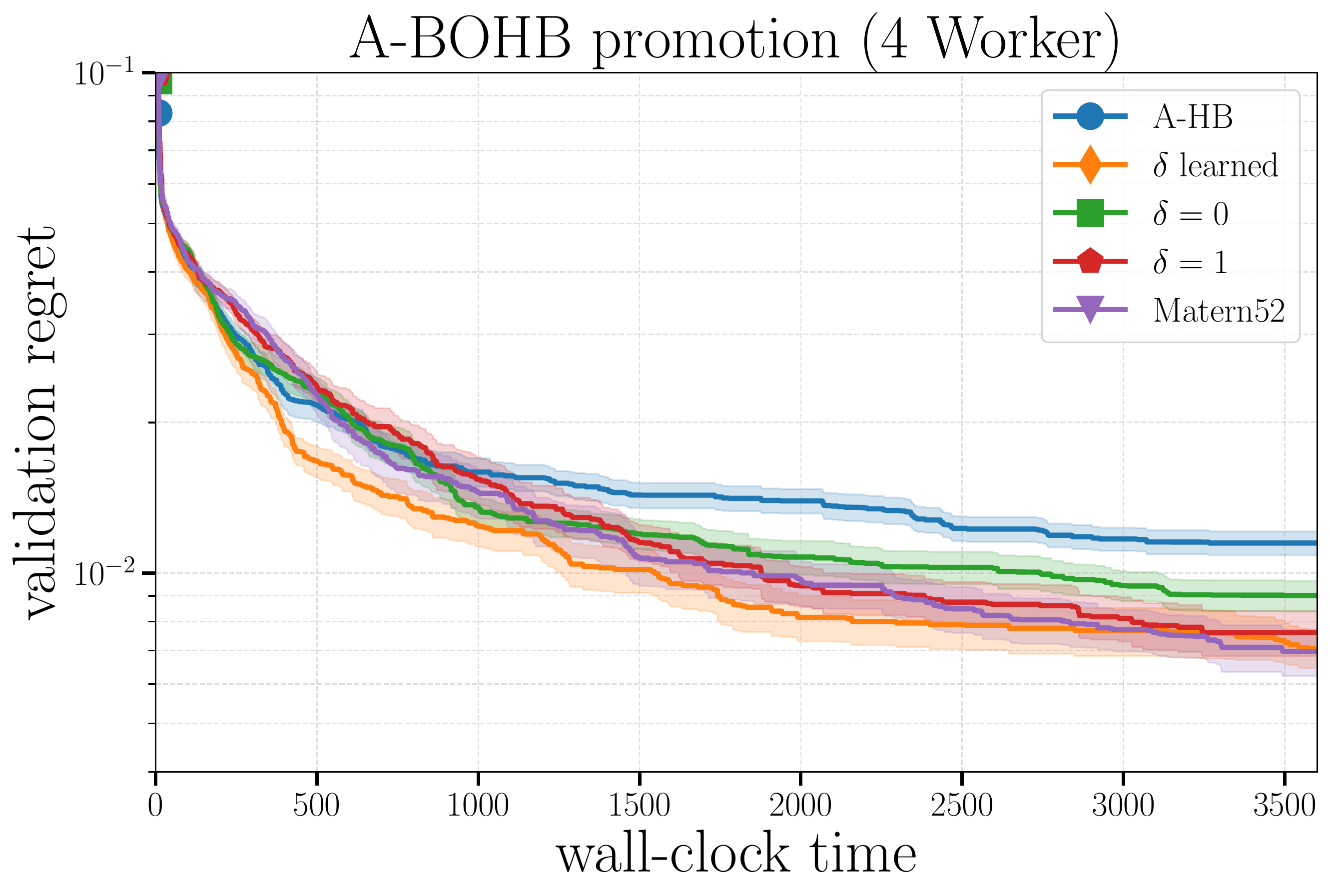}
 \includegraphics[width=.32\linewidth]{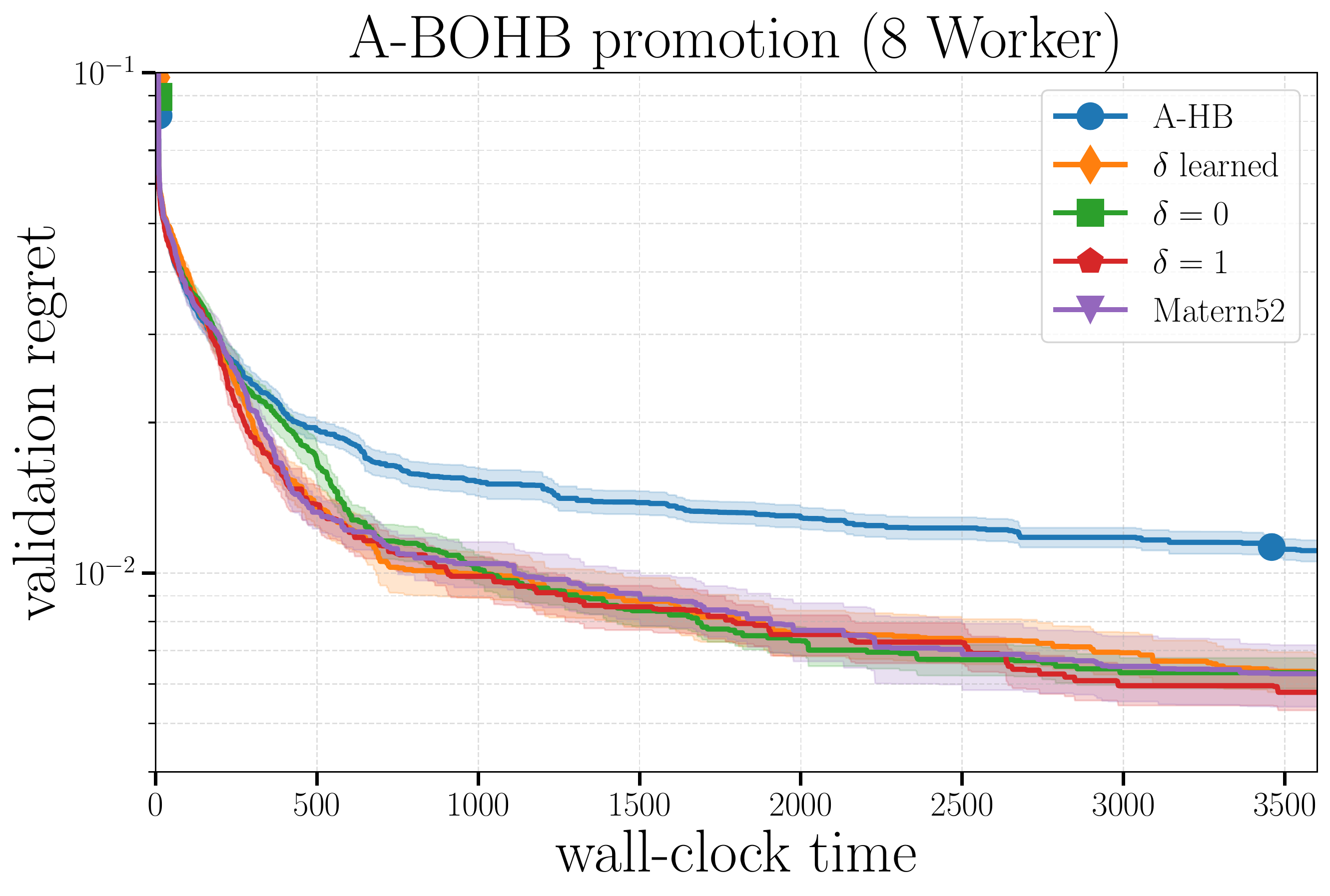}\\

 \caption{Effect of surrogate model (kernel, mean function) on performance of asynchronous BOHB on {\it electricity} dataset (top: stopping variant; bottom: promotion variant). We also include A-HB (random choices) for comparison. There is not a consistent trend across the difference kernels (note the standard error of the mean). }
\label{fig:kernels}
\end{center}
\end{figure*}

We analyse the influence of the choice of kernel and mean function for $f(\mbx, r)$ on the performance of the stopping and promotion variant of asynchronous BOHB. We compare different variants of the exponential decay model described in Section \ref{app:surrogate_models}: fixing $\delta=0$ as in Swersky et al.~\citep{Swersky:14}, $\delta=1$, or learning $\delta$ with the other GP hyperparameters. We use a Mat\'{e}rn $\frac{5}2$ ARD kernel for the $k_{\mathcal{X}}$ model between configurations $\mbx$ \cite{Snoek:12}. As an additional baseline, we use a Mat\'{e}rn $\frac{5}2$ ARD kernel and constant mean function on inputs $(\mbx, r)$.

In \figref{kernels} we compare the various types of kernel for different number of workers. Surprisingly, with two, four and eight workers there is virtually no difference between the surrogate models. While model-based decisions clearly matter (asynchronous BOHB outperforms asynchronous HB see main text), the precise model for inter- or extra\-polation along $r$ seems less important. For the experiments in the main paper, we use the exponential decay kernel with learned $\delta$ as the most flexible choice.


\fi
\end{document}